\documentclass[journal]{IEEEtran}
\usepackage{graphicx}
\usepackage{paralist}
\usepackage[inkscapelatex=false]{svg}
\usepackage[shortlabels]{enumitem}
\usepackage{hyperref}
\usepackage{multirow}
\usepackage{dblfloatfix} 

\usepackage{amsmath}
\usepackage{mathptmx}

\usepackage{algorithm}
\usepackage[noend]{algpseudocode}
\usepackage{algcompatible}
\usepackage{lineno}
\usepackage{comment}
\usepackage{amssymb}
\usepackage{microtype}
\usepackage{bm}
\usepackage{colortbl}
\usepackage{cite}

\begin{document}
\title{Exploiting Information Theory for Intuitive Robot Programming of Manual Activities}

\author{Elena~Merlo, Marta~Lagomarsino, Edoardo~Lamon, Arash~Ajoudani \vspace{-0.6cm}
\thanks{This work was supported by the Horizon Europe Project TORNADO (GA 101189557) and by the European Union under NextGenerationEU (FAIR - Future AI Research - PE00000013).}
\thanks{Elena~Merlo is with Human-Robot Interfaces and Interaction, Istituto Italiano di Tecnologia, Genoa, Italy and Dept. of Informatics, Bioengineering, Robotics, and Systems Engineering, University of Genoa, Genoa, Italy (e-mail: elena.merlo@iit.it). \textit{(Corresponding author: Elena Merlo.)}}
\thanks{Marta~Lagomarsino and Arash~Ajoudani are with Human-Robot Interfaces and Interaction, Istituto Italiano di Tecnologia, Genoa, Italy (e-mail: marta.lagomarsino@iit.it, arash.ajoudani@iit.it).}
\thanks{Edoardo~Lamon is with Dept. of Information Engineering and Computer Science, University of Trento, Trento, Italy (e-mail: edoardo.lamon@unitn.it).}}

\maketitle

\begin{abstract}
Observational learning is a promising approach to enable people without expertise in programming to transfer skills to robots in a user-friendly manner, since it mirrors how humans learn new behaviors by observing others. Many existing methods focus on instructing robots to mimic human trajectories, but motion-level strategies often pose challenges in skills generalization across diverse environments. 
This paper proposes a novel framework that allows robots to achieve a \textit{higher-level} understanding of human-demonstrated manual tasks recorded in RGB videos. By recognizing the task structure and goals, robots generalize what observed to unseen scenarios. 
We found our task representation on Shannon's Information Theory (IT), which is applied for the first time to manual tasks. IT helps extract the active scene elements and quantify the information shared between hands and objects. 
We exploit scene graph properties to encode the extracted interaction features in a compact structure and segment the demonstration into blocks, streamlining the generation of Behavior Trees for robot replicas. 
Experiments validated the effectiveness of IT to automatically generate robot execution plans from a single human demonstration. 
Additionally, we provide HANDSOME, an open-source dataset of HAND Skills demOnstrated by Multi-subjEcts, to promote further research and evaluation in this field.
\end{abstract}


\IEEEpeerreviewmaketitle

\vspace{-0.1cm}
\section{Introduction}

\IEEEPARstart{I}{n} recent decades, robotic platforms have gained considerable traction in industrial, healthcare, and domestic settings \cite{lorenzini2023ergonomic, sun2021robotic}. However, the challenges and technical expertise required for their programming have limited widespread accessibility and adoption. 
Consequently, many companies and research groups are focusing on developing intuitive programming methods and interfaces to streamline robot instruction processes, particularly for non-experts \cite{franka, abb, artiminds, steinmetz2018razer}.

Programming by Demonstration (PbD) simplifies robot programming by transferring skills via practical demonstrations \cite{billard2008survey, ravichandar2020recent, argall2009survey}, taking cues from human social learning processes such as emulation \cite{whiten2009emulation}.
In Kinesthetic Teaching (KT), also known as direct teaching, robots can record through their sensors and learn trajectories demonstrated by a user who manually guides them through haptic interaction within their workspace \cite{calinon2007learning}. 
Demonstrated movements can be segmented to establish a repository of fundamental motions that can be combined to produce more complex behaviors \cite{kulic2012incremental, takano2016real}. 
However, this method primarily finds application with articulated and lightweight robots, such as collaborative robots \cite{villani2018survey}. Moreover, although it simplifies the correspondence problem \cite{nehaniv2002correspondence}, inverse kinematics struggles with highly dynamic movements. This is because it becomes challenging to control all the robot's joints simultaneously in such scenarios. 

To avoid this issue, robots can alternatively observe an external demonstrator performing a task. This approach, denoted as Observational Learning (OL) \cite{pauly2021seeing}, necessitates the usage of accurate motion capture systems, which typically involve outfitting both the human demonstrator and objects in the environment with sensors. In \cite{terlemez2014master}, the authors introduced a framework that exploited motion data from a motion capture system to drive a sophisticated $104$ DoF human model, enabling the representation and reproduction of human motions in humanoid robotics. 
Gaussian Processes were instead investigated in \cite{arduengo2023gaussian} to effectively learn manipulation skills demonstrated by a human wearing the XSens suit. 
To prevent sensors from hindering humans during demonstrations, some researchers opted to capture human motion solely through processing data from RGB cameras \cite{desmarais2021review, fortini2023markerless, albrecht2011imitating}.
Despite its potential for human activity recognition \cite{beddiar2020vision, jegham2020vision}, RGB tracking still has limitations and is not entirely reliable for implementing a low-level PbD.

\begin{figure*}[t!]
\centering
\includegraphics[width=\linewidth]{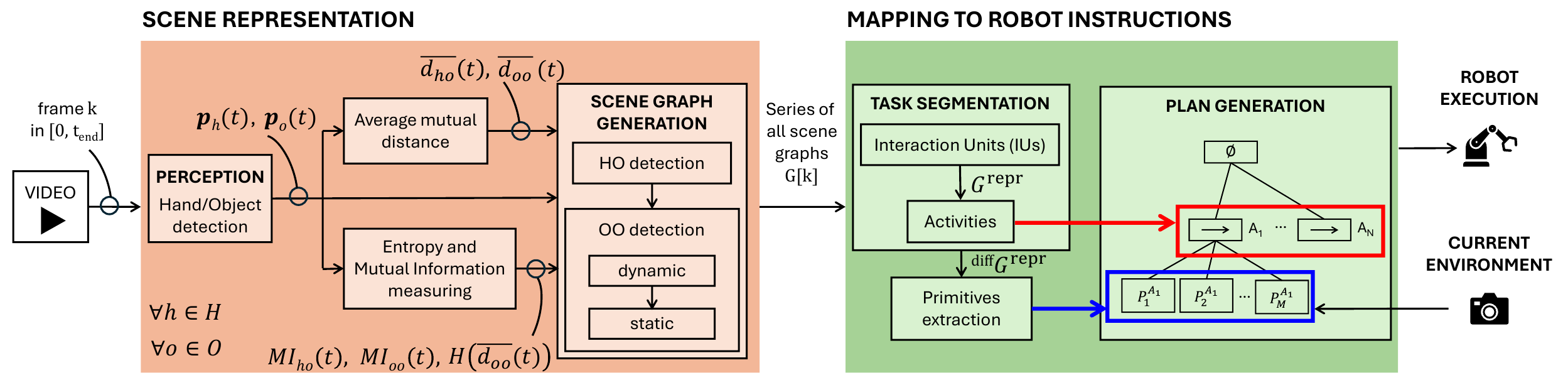}
\vspace{-0.7cm}
\caption{The overall structure of the proposed framework is composed of two main blocks:
(i) scene representation, and (ii) automatic map to robot instructions.}
\label{fig:framework}
\vspace{-0.2cm}
\end{figure*}

Moreover, approaches that focus on learning the trajectory without considering environmental conditions may fail when the same skill needs to be performed in unseen scenarios.
To address this challenge, researchers have investigated methods for segmenting demonstration trajectories into multi-steps to identify and extract reusable skills, utilizing inverse reinforcement learning techniques \cite{krishnan2019swirl, ranchod2015nonparametric}.
However, these methods present a strong potential but are still in the early stages and not yet aligned with the mission of intuitive programming. 

More practical methods that focus on creating \textit{high-level} representations of demonstrated tasks, enabling robots to generalize and abstract skills from a single human demonstration for replication, have been presented in the literature. Despite growing interest, this is still a relatively new area with many unresolved challenges.
Few studies embed the semantic representation of a skill demonstrated through KT by identifying key elements within the workspace that define the skill itself. 
In \cite{steinmetz2019intuitive}, an algorithm called Semantic Skill Recognizer was implemented to classify the demonstrated trajectory into a proper skill within a predefined set.
On the other hand, the authors of \cite{abdo2013learning} extracted and monitored object features to retrieve skill preconditions (e.g., the fact that an object must be grasped before placing it on top of another object) and effects (e.g., the relative position of the two objects at the end of the motion). 
The compact description of object relations through dynamic Semantic Networks, i.e., a graph-based representation constantly updated throughout the demonstration, was exploited in \cite{zanchettin2023symbolic}. In this work, changes in the graph structure alerted for variations that the performed actions caused in the environment. 
Understanding such mapping enables the formulation of robot execution plans through the Planning Domain Definition Language (PDDL) \cite{aeronautiques1998pddl}, which can also be automatically generated after a few demonstrations \cite{zanchettin2023symbolic, diehl2021automated, abdo2013learning}. 
To support semantic capture during KT mode, the authors of \cite{stenmark2018supporting} required semantic annotations through natural language by the programmer to ground the collected motion data. The KT-based strategy proposed by \cite{caccavale2019kinesthetic} was instead combined with an attentional system that supervised the demonstration and related the segmented basic movements to a known task structure. In \cite{chou2022learning}, the robot learns a linear temporal logic (LTL) formula based on semantic atomic propositions that outlines the sequence in which the sub-tasks should be executed. Such a formula is then used to determine the optimal trajectories to perform in the current scenario.

The task demonstration by moving the robot still remains less intuitive than directly performing it and enabling robot imitation, especially for non-experts. For this reason, studies have focused on analyzing human movements and the properties of the objects in the environment with which humans interact to derive mappings between movement and its meaning automatically \cite{guha2013minimalist, ramirezamaro2014automatic, memmesheimer2020robotic, ramirezamaro2017transferring}. The inferred human actions are then mapped into a set of predefined robot primitives. Yet, these model-based methods imply formulating lists of rules for recognizing each specific action \cite{ziaeetabar2018recognition}, making it challenging to consider and adequately describe all possible scenarios. 
Other approaches aimed to understand the observed action by inspecting the sequence of changes within the scene throughout the entire task execution. Authors of \cite{aksoy2011learning, savarimuthu2017teaching, aein2019library} utilized Semantic Event Chains (SECs), i.e., a matrix encoding changes of spatial and motion relationships among scene elements across the entire task execution, and learned how to associate specific actions with the corresponding SEC.
Alternatively, the task can be segmented based on scene changes, and isolated actions can be identified by consulting the Object-Action Complex library \cite{wachter2013action}. This library includes mappings to actions according to their preconditions and predicted effects on the world state.
Knowing such mappings, if environment conditions are favorable \cite{kroemer2016learning}, the robot can autonomously plan and execute manipulations aimed at transitioning the world from the initial state to the goal state \cite{zeng2018semantic, sui2017goal}.
Recent advancements include machine learning-based techniques utilizing specialized neural networks to process RGB image sequences. In this case, natural language descriptions of video content are generated and then segmented to extract the sequence of commands for the robot's execution \cite{nguyen2018translating, yang2023watch}. Similarly, Graph Neural Networks (GNNs) have been employed to segment and recognize human actions using graph-based representations \cite{dreher2020learning}. Using GNNs, authors of \cite{zhu2021hierarchical} generated task and motion plans, leveraging a two-level representation integrating both symbolic and geometric scene graphs.
Nevertheless, these methods are often time-consuming and may require a large amount of data.

This work embraces the intuitiveness of programming by simply recording demonstrations through RGB cameras. Moreover, it introduces a novel and cost-effective one-shot technique for teaching robots to perform manipulation tasks using only one human demonstration. 
The proposed method automatically segments the video into actions, identifies a sequence of motion primitives, and generates a plan for robot execution, translating human actions into robot instructions.
Unlike existing approaches, we do not solely rely on spatial data in task analysis and representation but integrate measures of uncertainty and information content, facilitating the generalization of robots' execution plan to various environmental conditions. Central to our method is Shannon's Information Theory (IT) \cite{bossomaier2016introduction}, which, to the best of our knowledge, is applied for the first time to describe and interpret manipulation tasks, uncovering how entropy measures can highlight the active components of the scene and quantify the information sharing and exchange between hands and objects.

The overall structure of our framework is represented in Fig. \ref{fig:framework}. The framework is composed of two main blocks:
\begin{inparaenum}[(i)]
    \item scene representation, and
    \item mapping to robot instructions.
\end{inparaenum}
The initial block processes each image in the video demonstration, converting it into a graph structure. This graph encodes the instantaneous relations between hands and objects, incorporating their detected poses from the perception module, their mutual distances, and the information they convey and share based on measures from IT.
The second block examines the complete series of graphs over time, utilizing their properties to segment the task into distinct activities. For each activity, a set of motion primitives is extracted by analyzing preconditions and effects of all the interactions that constitute it. The retrieved hierarchical task representation is mapped to a Behavior Tree (BT) plan for robot execution \cite{colledanchise2018behavior, fusaro2021human}, whose structure remains unchanged once generated. However, the plan adjusts to the environmental conditions during the robot replica. The framework has been tested on multi-subject demonstrations and compared with state-of-the-art methods relying on position and velocity evaluations.

The rest of this article is structured as follows. In Section II, we detailed further the functioning of the scene representation process, with a special focus on how we used IT to describe manipulation tasks, while Section III is devoted to explain how we map human executions to robot language. Section IV describes experiments conducted to both test the validity of the task representation and the effectiveness of the mapping to robot instructions. Section V covers the experimental results, discussed in Section VI. Section VII provides the conclusions.
\vspace{-0.5cm}

\section{Scene Representation}
This section presents the processing of each frame in the task demonstration and details the utilization of entropy analysis within video clips to represent the task via a time series of scene graphs. 
A \emph{perception} module identifies hands and objects in each frame $k$ taken at time instant $t$, capturing the 6D pose $\mathbf{p}_{h}{\scriptstyle (t)}$ of each hand $h \in H$ and the pose $\mathbf{p}_{o}{\scriptstyle (t)}$ of each object $o \in O$ in the scene. Subsequently, the positional data are sent to the entropy calculation module. This module evaluates the information conveyed by each position signal and the information exchanged between pairs of scene elements, focusing on time window $w$ centred in $t$. The position signals are also processed to obtain the average distances between hands and objects, computed again using $w$. 
These outputs are then utilized to construct a \emph{scene graph} structure $G[k]$ \cite{chang2023comprehensive}, that encodes relationships among scene elements. 
Specifically, $G[k]$ nodes represent hands and objects detected within frame $k$, while edges between nodes evidence the presence of interactions between the connected elements. 
All the generated graphs are finally stored.

Since, to the best of our knowledge, it is the first time that IT is exploited in manual task representation, the initial subsection is dedicated to explaining the motivation behind its usage, the metrics involved, and the methodology for their computation.

\subsection{Information Theory in Manipulation Tasks}

To achieve an effective task representation, it is essential to identify the \textit{active} part of the scene, namely the area where significant changes occur \cite{kuniyoshi1994learning}.
This enables us to create a more concise description and focus the task comprehension solely on the elements relevant to executing the task while filtering out the surrounding background elements. 
Shannon's IT emerges as a potent tool for this purpose. Indeed, IT enables quantifying the amount of information conveyed by a signal, without considering its semantic meaning. Specifically, the key measure of the information content is \emph{entropy}, representing the average level of uncertainty inherent in the signal and surprise associated with its possible values. 
Given a random variable $X$ linked to an event space $\Omega_x$, the Shannon's entropy of $X$ is given by:
\begin{equation}
\label{eq:entropy}
    H(X) = - \sum_{x \in \Omega_x} p(x) \cdot \log_{2}{p(x)},
\end{equation}
where $x$ is a discrete measurement from $\Omega_x$ and $p(x)$ denotes the probability of such a measurement. 
Note that Shannon's entropy is measured in bits. The higher the entropy value, the more bits are required to transmit the information contained in the signal. 

In our framework, we primarily utilize entropy on position signals to understand scene dynamics. 
However, instead of computing the entropy for the entire signal at once, we use a sliding time window $w$ and calculate entropy values as the window moves along the signal. That is, we apply Eq. (\ref{eq:entropy}) at each shift of $w$, obtaining a time series of entropy values $H(X{\scriptstyle (t)})$. By analyzing the trend of this new signal, we can explore how and when entropy changes during task execution.
For each position of the window centered at $t^*$:
\begin{inparaenum}[]
    \item $X$ represents the random variable corresponding to the scene element's positional signal, empirically described by the set of samples $[x{\scriptstyle (t^*-w/2)}, \dots, x{\scriptstyle (t^*+w/2)}]$;
    \item $\Omega_x$ is the set $\{x{\scriptstyle (t^*-w/2)}, \dots, x{\scriptstyle (t^*+w/2)}\}$;
    \item $x$ is each single sample in $\Omega_x$;
    \item $p$ is the probability distribution for $X$ derived from the empirical measurements in $w$;
    \item and $H(X{\scriptstyle (t^*)})$ is the resulting entropy value.
\end{inparaenum}

The probability distribution $p$ is represented as a histogram, where each bin captures the frequency of occurrence of specific values within the current time window. These values are grouped based on quantization intervals, which define the bin widths. The height of each bin corresponds to how often values fall within that interval. In our approach, we chose a fixed quantization interval $q$, resulting in a variable number of bins depending on the motion range. 
The rationale for computing the probability distribution and entropy over the trajectory within a time window $w$ is to detect variations in motion patterns by examining the distribution of positional data.
For instance, when an object moves in a straight line at a constant speed, the entropy signal will be constant, reflecting steady positional variability over the shifting $w$. However, if the object begins to decelerate, its position values become more predictable as it approaches a stop, leading to a decrease in entropy.

By fixing $q$, we can distinguish between movements confined to a small space and those covering a broader range. For movements within a restricted area, as $w$ shifts, $p$ will have fewer bins with higher probabilities, as the moving item frequently assumes similar or identical positions, resulting in low entropy. Conversely, for broader movements, $p$ will feature more bins due to the diversity of positions covered, potentially resulting in a flat distribution and high entropy.
Moreover, if we select a size for $w$ that can include more repetitions of the same movement, multiple samples will might have the same value or be similar enough to fall within the same $q$. 
This might be the case of stirring a cup with a spoon and then placing the spoon on a spot on the table, several centimeters away from the cup. During the stirring motion, if $w$ is wide enough to cover the repetitive nature of the task, the position signal will probably take on similar values cyclically, keeping $H{\scriptstyle (t)}$ at a low value. In contrast, when the spoon is moved to be placed on the table, the trajectory may become broader, covering different points. As a result, we would likely observe a peak in $H{\scriptstyle (t)}$ during the spoon's relocation.

\begin{figure} [t!]
\centering
\includegraphics[scale=0.55]{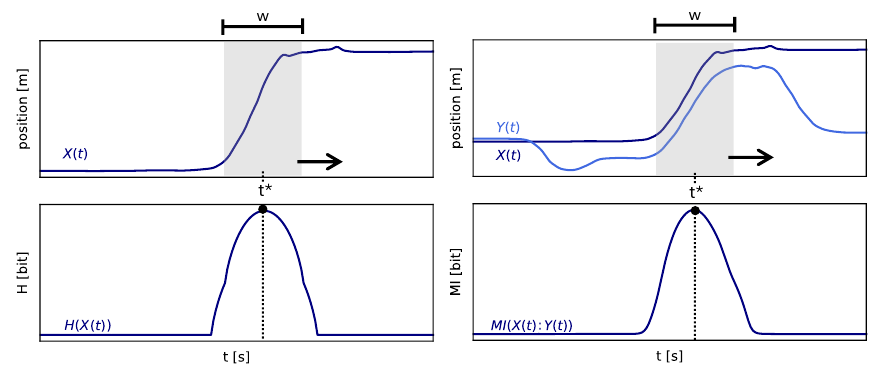} 
\vspace{-0.4cm}
\caption{(left) The scenario of an object being relocated is presented. $X{\scriptstyle (t)}$ represents its $1$D position signal over time. $H(X{\scriptstyle (t)})$ is computed by shifting $w$ over $X{\scriptstyle (t)}$ and computing entropy at each time point considering samples included by $w$. The bell-shaped curve flags the position variation. (right) The case where a hand moves an object is depicted. The $1$D position signal of the hand is denoted as $Y{\scriptstyle (t)}$ and that of the object as $X{\scriptstyle (t)}$. By computing $MI(X:Y)$ at each time point while shifting the window $w$, a bell-shaped curve is obtained corresponding to when the hand and the object move together.}
\label{fig:H(t)}
\vspace{-0.3cm}
\end{figure}

What is noteworthy is that by analyzing the entropy signal, we can detect changes in the motion pattern just based on the distribution and variability of positional data. This method offers a simpler way to identify significant motion changes, bypassing the need for analyses of velocity or acceleration profiles, which can often be prone to errors from numerical estimations.
As an illustration, Fig. \ref{fig:H(t)} (left) depicts the $1$D position signal $X{\scriptstyle (t)}$ of an object being relocated. By sliding the window $w$ along the signal $X{\scriptstyle (t)}$ and computing the entropy at each instant, considering the values within $w$, we observe a bell-shaped temporal trend of the entropy $H(X{\scriptstyle (t)})$. The peak at $t^*$ indicates the central moment of the variation, while the amplitude and duration of the bell-shaped curve offer insights into its speed and magnitude. 
Additionally, analyzing the time derivative at each time point $\frac{dH(X{\scriptstyle (t)})}{dt}$ provides understanding of when the object initiates movement ($\frac{dH(X{\scriptstyle (t)})}{dt} > 0$) and when it is reaching a new equilibrium ($\frac{dH(X{\scriptstyle (t)})}{dt} < 0$).

However, to describe manual tasks, we need to go beyond analyzing individual scene elements in isolation and examine their interactions. 
The term \emph{interaction} refers to a situation where two or more entities communicate with each other or react to each other. This means that the elements involved in the interaction engage in a mutual exchange of information. Once again, Shannon's entropy proves effective for estimating the information shared among multiple signals and thus for examining interactions. 
To quantify this shared information, we rely on \emph{mutual information} ($MI$), a metric that measures the statistical dependence between the variables involved. 
If we visualize the entropy of two distinct random processes $X$ and $Y$ as two sets $H(X)$ and $H(Y)$, the $MI$ corresponds to their intersection and is calculated as follows:
\begin{equation}
MI(X:Y) = H(X) + H(Y) - H(X,Y), 
\label{eq:mi}
\end{equation}
where $H(X,Y)$ is called \emph{joint entropy} and represents the union of the two sets:
\begin{equation}
\label{eq:joint_e}
H(X,Y) = - \sum_{x \in \Omega_X} \sum_{y \in \Omega_y} p(x,y) \cdot \log_{2}{p(x, y)},
\end{equation}
where $x$ and $y$ are discrete measurements from $\Omega_x$ and $\Omega_y$, respectively, and $p(x,y)$ is the joint probability of both events occurring simultaneously in the time window $w$.
Eq. (\ref{eq:mi}) implies that, if $MI(X:Y) = 0$, $X$ is independent of $Y$ and vice versa, due to the symmetry of this measure. In other words, knowledge about the value of one variable provides no information about the other variable. Since we are interested in observing the dynamics of $MI$ over time, we apply the same approach described earlier, shifting the time window $w$ along the considered signals to obtain $MI(X{\scriptstyle (t)}:Y{\scriptstyle (t)})$.

By applying $MI$ to the position signals of two objects, we can determine whether their movements are dependent or independent. In the context of manual activities, this allows us to reliably establish whether two scene elements, such as a hand and a manipulated object, are jointly moving and to monitor changes in this movement over time, as $MI$ relies on the entropy of the two position signals. Moreover, $MI$ can capture both linear and nonlinear relationships between signals, facilitating the recognition of more complex dependencies that may not be evident by observing the velocity profile of the involved bodies. For instance, when a hand carries multiple objects on a tray, $MI$ can identify the hand and objects moving as a unified unity, even if objects may slide or oscillate during transportation. 
In Fig. \ref{fig:H(t)} (right), we illustrate the simple case where a hand grasps an object and moves it. We consider the $1$D position signal of the hand, denoted as $Y{\scriptstyle (t)}$, and that of the object, denoted as $X{\scriptstyle (t)}$. By computing $MI(X:Y)$ at each time point while shifting the window $w$, we obtain a bell-shaped curve corresponding to the coordinated movement of the hand and the object. During the approach and retreat phases, $MI$ is zero. This strengthens the definition of interaction between hand and object, especially in cluttered environments where mere proximity may not imply interaction, such as when the hand approaches an object to reach another nearby.
Note that in our framework, we study the movements and interactions within $3$D space, hence when evaluating the $MI$ value between a hand $h \in H$ and an object $o_i \in O$, denoted as $MI_{ho_i}$, we sum the $MI$ calculated across all three spatial directions.

\vspace{-0.2cm}
\subsection{Scene Graph Generation}
A scene graph $G$ is a directed graph data structure, which can be formally expressed as a tuple $G = (V, R, E)$, where $V$ is the set of nodes, $R$ represents a set of relationships between the nodes, and $E$ denotes the edges connecting the nodes. Each node is expressed as $v_\text{i} = (c_\text{i}, \hspace{1mm} A_\text{i})$, where $c_\text{i}$ and $A_\text{i}$ respectively indicate its identity and attributes~\cite{chang2023comprehensive}. 
In our framework, $G$ nodes indicate hands and objects detected within the scene. Their only attribute is their pose $\mathbf{p}_i \in \mathbb{R}^6$. 
Edges between nodes define interactions between connected elements whose features are described by set $R$. 
In uni-manual activities, we categorize interactions into two main groups:
\begin{itemize}
    \item \emph{hand-object interactions}, occurring between a hand and an object,
    \item \emph{object-object interactions}, occurring between different objects and further divided into subcategories. 
\end{itemize}
If an interaction between two elements is found, an edge between the two corresponding nodes is established. 

The process of scene graph generation starts by searching for hand-object interactions since the hand is the element that can generate changes within the environment. Then, object-object interactions are analyzed. If the first phase does not detect any hand-object interactions, the second phase is not addressed.  

Let us now analyze the scene graph generation procedure adopted for a single hand $h \in H$ at frame $k$. Note that the use of IT has greatly simplified interaction detection and identification, making the process intuitive and easily programmable. Further details will be provided in the subsequent subsections.

\subsubsection{Hand-Object Interactions Detection}
We delineate two types of Hand-Object interactions ($HO$): \textit{manipulation} and \textit{contact-only}.
The former denotes scenarios where the hand holds an object and moves it.
In contrast, a \textit{contact-only HO} occurs when the hand maintains contact with the object without any accompanying movement. 

\begin{figure}[h!]
    \vspace{-0.2cm}
    \centering
    \includegraphics[width=\linewidth]{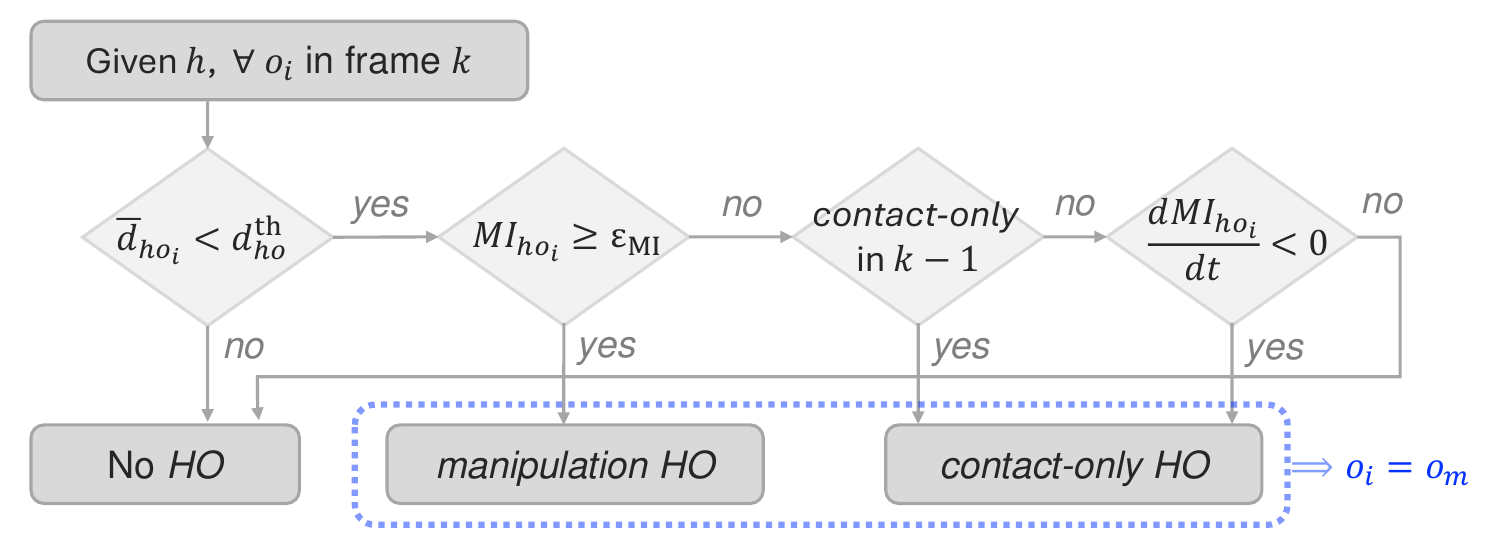}
    \vspace{-0.8cm}
    \caption{Detection of Hand-Object interactions ($HO$), categorized into \textit{manipulation} and \textit{contact-only} $HO$.}
    \label{fig:flowchart_HO_int}
\end{figure}

In the flowchart displayed in Fig. \ref{fig:flowchart_HO_int}, the procedure to detect and define the interaction between a hand $h$ and a generic object $o_i \in O$ within the scene in frame $k$ is illustrated. 
For each $HO$ to occur, a necessary condition is that the elements involved must be sufficiently close to each other. Therefore, their average distance $\overline{d}_{ho_i}$ must be under a given threshold $d^{th}_{ho}$; otherwise, no interaction exists.
A \textit{manipulation} $HO$ is generated when the mutual information $MI_{ho_i}$ between $h$ and $o_i$ is greater than a threshold $\epsilon_{MI}$, which is a value very close to 0, indicating that the object $o_i$ is manipulated. 
Contrarily, if $MI$ is decreasing and small enough (i.e., $\frac{dMI_{ho_i}}{dt}<0$ and $MI_{ho_i} < \epsilon_{MI}$) or if a \textit{contact-only} $HO$ has already been detected for frame $k-1$, a \textit{contact-only} $HO$ is established or confirmed. 
As soon as the average distance exceeds the minimum hand-object distance, no more interaction exists.

Note that, according to our formulation, \textit{contact-only} $HO$ can occur solely at the conclusion of the manipulation process, i.e., after a \textit{manipulation} $HO$ but not before it. This design ensures that the algorithm does not recognize any interactions unless the hand has manipulated an object. This adds robustness in handling tasks performed in cluttered environments.
The operations are repeated for each object $o_i \in O$ detected within the scene, starting from the closest to the furthest with respect to $h$. Upon detecting one $HO$ interaction, the algorithm proceeds directly to detecting object-object interactions.

\subsubsection{Object-Object Interactions Detection}
Initially, Object-Object interactions ($OO$) are categorized into \textit{dynamic} and \textit{static} types. This classification depends on the behavior of the object (i.e., a generic object $o_j \in O$ in the scene) interacting with the one in hand (i.e., the manipulated object denoted as $o_m \in O$). \textit{Dynamic} $OO$ occur when the object $o_j$ is in motion, while \textit{static} $OO$ happen when $o_j$ is stationary.

\textit{(2a)} \hspace{0.01cm} \textit{Dynamic OO}:
A \textit{dynamic} $OO$ is established if and only if the involved objects move cohesively, as a unity.

\begin{figure}[h!]
    \centering
    \includegraphics[width=\linewidth]{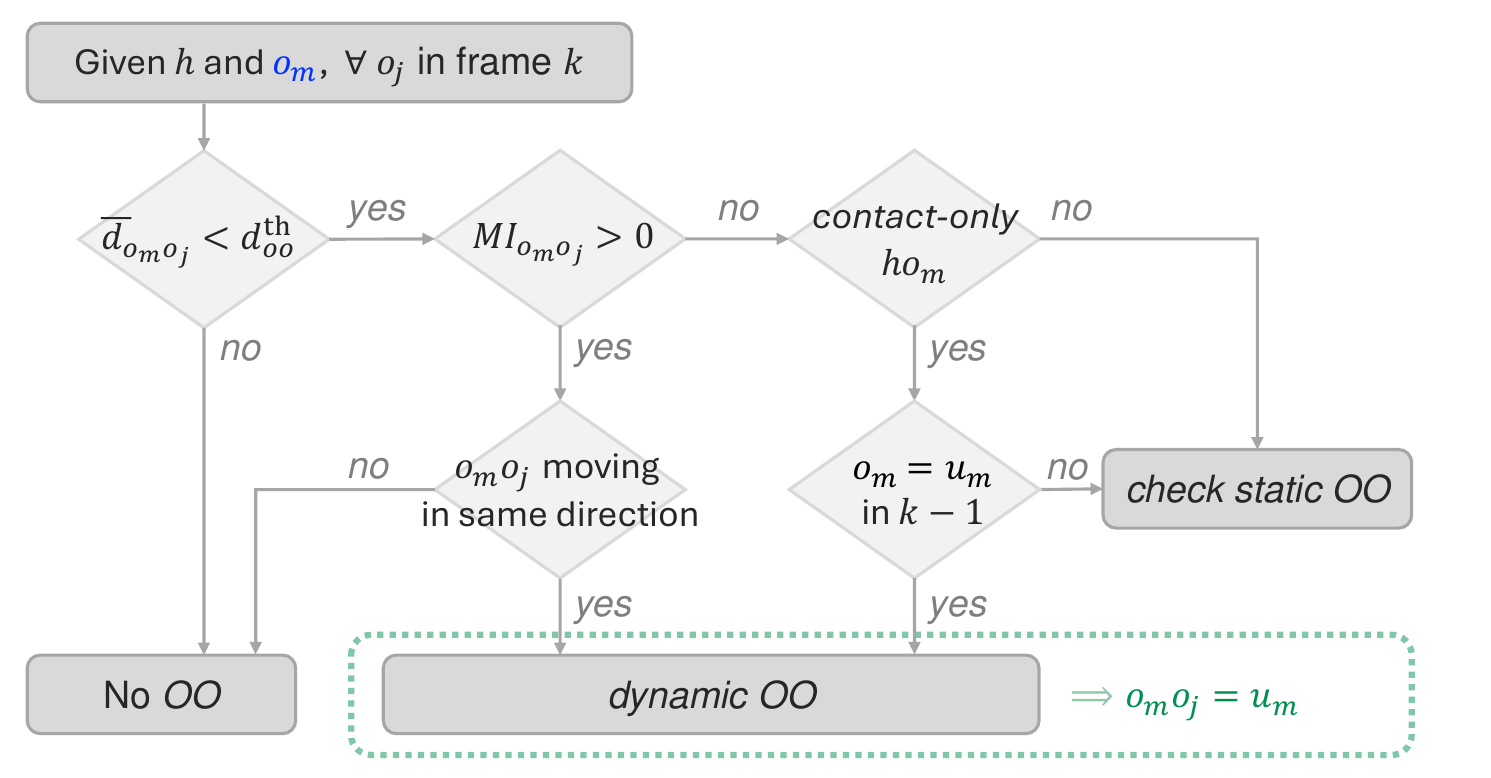}
    \vspace{-0.7cm}
    \caption{Detection of \textit{dynamic} Object-Object interactions ($OO$). If conditions are satisfied, the manipulated object $o_m$ and a generic object $o_j$ within the scene are considered a manipulated unity $u_m$.}
    \label{fig:flowchart_dynamicOO_int}
    \vspace{-0.05cm}
\end{figure}

As illustrated in the flowchart of Fig. \ref{fig:flowchart_dynamicOO_int}, the manipulated object $o_m$ and a generic object $o_j$ in frame $k$ are considered part of a moving unity if they are close enough (i.e., $\overline{d}_{o_m o_j} < d^{th}_{oo}$), share information ($MI_{o_m o_j} > 0$), and move in the same direction.
To determine if the two objects are moving in the same direction, we compute the angle between their displacement vectors, considering the positions assumed at $t$ and at $t + w/2$. 

These operations are repeated until all the objects in the scene are examined. To check if multiple objects are part of the same manipulated unity, the process continues by measuring co-information, one generalization of $MI_{oo}$ to multiple variables.
At the end of this step, if \textit{dynamic} $OO$ interactions have been detected, the manipulated object $o_m$ becomes a manipulated unity of objects $u_m$. 
If a \textit{contact-only} $HO$ occurs between the hand and one of the objects in $u_m$, this extends to a \textit{contact-only} interaction with the entire $u_m$. However, once the hand moves away, the objects within the unity are once again treated as separate entities.
From now on, with $o_m$ we refer to both $o_m$ and $u_m$.

\begin{figure}[h!]
    \centering
    \includegraphics[width=\linewidth]{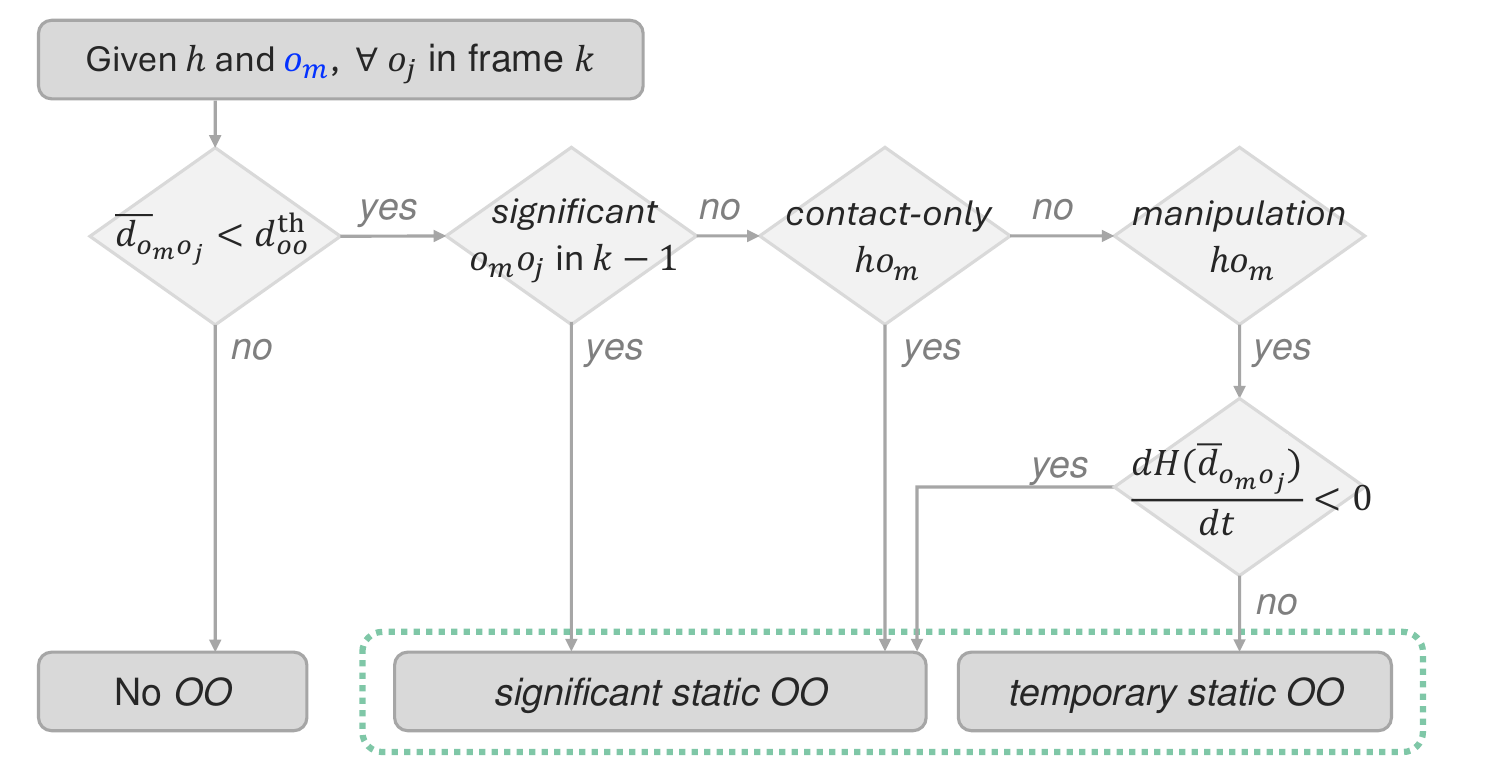}
    \vspace{-0.7cm}
    \caption{Detection of \textit{static} Object-Object interactions ($OO$). According to the trend of the entropy of objects' average distance $H(\overline{d}_{o_m, o_j})$, $OO$ can be either \textit{significant} or \textit{temporary}.}
    \label{fig:flowchart_staticOO_int}
    \vspace{-0.2cm}
\end{figure}

\textit{(2b)} \hspace{0.01cm} \textit{Static OO}: 
\textit{Static} $OO$ involve the in-hand object $o_m$ and a stationary background object $o_j \in O$. 
These interactions are categorized into two types: \textit{significant} and \textit{temporary}. This categorization discerns between interactions that are certain and indicative of a specific activity (i.e., \textit{significant}) from those occurring in isolated instances, which might not happen in subsequent repetitions, being not representative of such activity (i.e., \textit{temporary}). 
This differentiation proves effective during the plan generation phase, where only \textit{significant} $OO$ will be treated for mapping into robot commands, while \textit{temporary} ones will be disregarded (see Section \ref{sec:mapping}).
In Fig. \ref{fig:flowchart_staticOO_int}, the procedure to detect and define \textit{static} $OO$ is illustrated. 
Note that \textit{temporary} $OO$ can also evolve into \textit{significant} $OO$ over time.

For each \textit{static} $OO$ to occur, the involved objects must be sufficiently close to each other ($\overline{d}_{o_m o_j} < d^{th}_{oo}$). Then, if there is a \textit{contact-only} $HO$ between $h$ and $o_m$, or if a \textit{significant} $OO$ between $o_m$ and $o_j$ has already been detected in frame $k-1$, a \textit{significant static} $OO$ is established or confirmed. As a result, if after manipulating an object, the hand remains stationary with the object $o_m$ near $o_j$, such proximity is deemed \textit{significant}.

On the contrary, if the hand is manipulating $o_m$ (i.e, a \textit{manipulation} $HO$ exists between $h$ and $o_m$), we examine the time derivative of $H(\overline{d}_{o_m o_j})$, which denotes the entropy of the average distance between $o_m$ and $o_j$. Note that the $MI$ cannot establish if two objects interact when one remains stationary, i.e., when the entropy of the position signal for $o_j$ is zero. Therefore, $H(\overline{d}_{o_m o_j})$ is exploited to gain insights into the magnitude and speed of average distance variations. Specifically, if $\frac{dH(\overline{d}_{o_m o_j})}{dt} < 0$, $OO$ is regarded as \textit{significant}; otherwise, it is considered \textit{temporary}. 
When we define a \textit{significant} $OO$, we recognize that the interaction between the involved objects is becoming stable, indicating a tendency towards a predictable state of interaction. For this reason, we are interested in observing the decreasing trend of $H(\overline{d}_{o_m o_j})$ as it indicates the approach toward a new state of equilibrium. 
At this point, even if $H(\overline{d}_{o_m o_j})$ starts to vary rapidly, the interaction persists \textit{significant}, unless $\overline{d}_{o_m o_j}$ exceeds $d^{th}_{oo}$.
On the other hand, if $H(\overline{d}_{o_m o_j})$ increases as soon as $\overline{d}_{o_m o_j}$ becomes lower than $d^{th}_{oo}$, it means that $OO$ is not reaching stability but is just transitory.
For clarity, let's consider the scenario where an object $o_1$ is grabbed near another object $o_2$ and then moved to the opposite side of the table. As soon as $o_1$ is grabbed and starts moving, a \textit{temporary} $OO$ between $o_1$ and $o_2$ occurs, because $\overline{d}_{o_1 o_2}$ is below $d^{th}_{oo}$, but their distance changes rapidly until they are no longer close enough.
The operations are repeated for every object $o_j$ detected within the scene, starting from the closest to the furthest with respect to $o_m$. Upon detecting one $OO$, the algorithm terminates. 
The requirement for a \textit{static} $OO$ to necessarily involve an object in-hand prevents the generation of additional interactions among multiple stationary objects in the background.
If the user interacts with only one specific background object, we will not observe any difference in the topology of the graph, whether the environment is controlled or cluttered. 
This ensures consistency in the representation, regardless of the environmental conditions.

\begin{figure}[!t]
\centering
\includegraphics[scale=0.3]{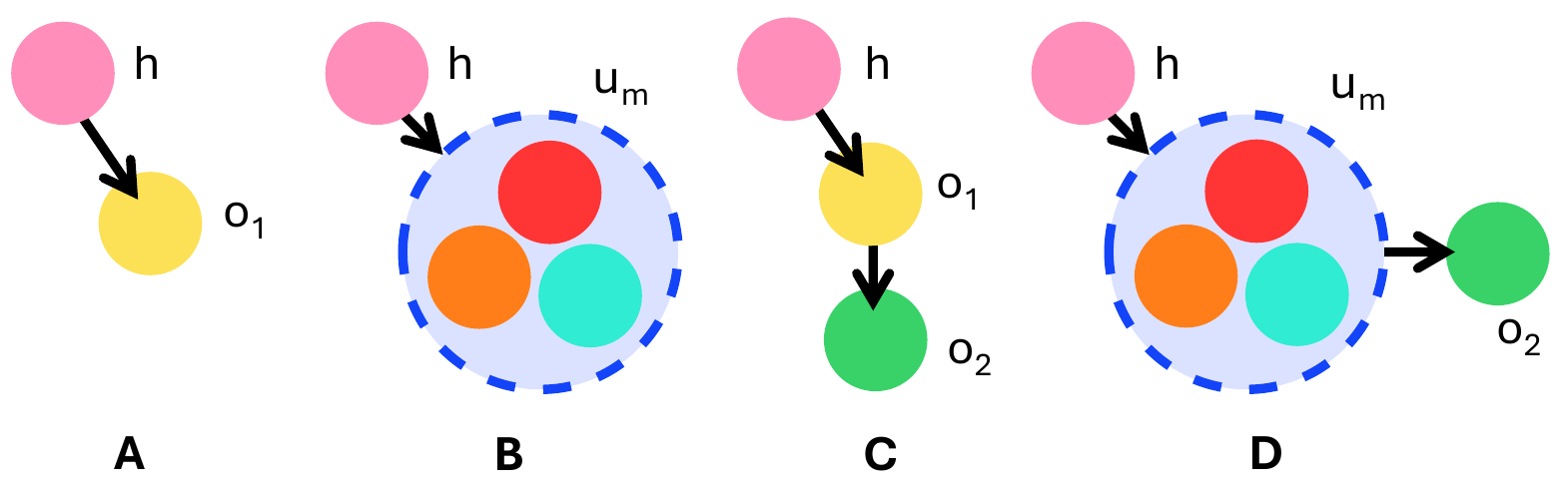}
\vspace{-0.2cm}
\caption{Topology of scene graph obtainable at the end of the interaction detection process: topology A reports only one $HO$ between the hand and the object $o_1$; topology B depicts \textit{dynamic} $OO$ retrieved for three objects whose nodes form the moving unity $u_m$; in topology C is encoded a static $OO$ between objects $o_1$ and $o_2$ as well; in topology D the manipulated unity $u_m$ interacting with the background object $o_2$ is reported.}
\vspace{-0.2cm}
\label{fig:graph_topologies}
\end{figure}

At the end of the interaction detection process, $G[k]$ assumes one topology among those depicted in Fig. \ref{fig:graph_topologies}.
If both $HO$ and \textit{static} $OO$ are retrieved, $G[k]$ obtained at instant $t$ is described by:
\begin{inparaenum}[]
    \item $V = \{h, o_1, o_2\}$, where $h = (\text{ID}_h, \mathbf{p}_h{\scriptstyle (t)})$, $o_1 = (\text{ID}_{o_1}, \mathbf{p}_{o_1}{\scriptstyle (t)})$, and $o_2 = (\text{ID}_{o_2}, \mathbf{p}_{o_2}{\scriptstyle (t)})$;
    \item $E = \{e_{h \rightarrow o_1}, e_{o_1 \rightarrow o_2}\}$, and
    \item $R = \{r_{h \rightarrow o_1}, r_{o_1 \rightarrow o_2}\}$, where $r_{h \rightarrow o_1}$ contains $MI_{ho_1}{\scriptstyle (t)}$ and the detected $HO$ type, and $r_{o_1 \rightarrow o_2}$ contains only the $OO$ type.
\end{inparaenum}
When a moving unity $u_m$ is involved, a new node is created to include all the objects within. As a result, instead of node $o_1$, the set $V$ contains node $u_m$, defined by the IDs of the objects within $u_m$ and the pose of the object in $u_m$ that is closest to $h$.
When a \textit{static} $OO$ is not identified, $G[k]$ is represented by $V = \{h, o_1\}$, $E = \{e_{h \rightarrow o_1}\}$, and $R = \{r_{h \rightarrow o_1}\}$.
The interaction types that we detect are summarized in Table \ref{tab:interactions}.

\begin{table}[!t]
\caption{Interaction types}
\label{tab:interactions}
\centering
\begin{tabular}{|c||c|c|}
\hline
 & \textbf{Type} & \textbf{Meaning}\\
\hline
\multirow{ 2}{*}{$HO$} & \textit{manipulation} & $h$ moves $o_m$\\
& \textit{contact-only} & $h$ just holds $o_m$\\
\hline
\multirow{ 3}{*}{$OO$} & \textit{dynamic} & $o_m$ and $o_j$ move jointly\\
& \textit{static significant} & $o_m$ persists close to stationary $o_j$\\
& \textit{static temporary} & $o_m$ passes by stationary $o_j$\\
\hline
\end{tabular}

\end{table}

\section{Mapping to Robot Instructions}
\label{sec:mapping}
At the end of the representation process, we collect a series of scene graphs, ordered over time.
By grouping similar scene graphs, we can identify atomic units termed \textit{Interaction Units} (IUs), corresponding to time segments where interacting scene elements remain unchanged. \textit{Activities} are retrieved by gathering time-ordered IUs that are logically connected.\footnote{The hierarchical description of manual tasks, which envisions a collection of activities formed by IUs as atomic blocks, is inspired by taxonomy in \cite{merlo2023automatic}.} The process of isolating each activity is called \emph{task segmentation}.
Subsequently, by examining the sequence of IUs within each retrieved activity, we can derive a sequence of robot motion primitives. 
This leads to a hierarchical representation of the task which fits properly with the robot plan structure given by Behavior Trees (BTs). 
Note that the procedure outlined below allows for the automatic conversion of movements demonstrated by a single hand $h \in H$ into a robot execution plan.

\subsection{Task Segmentation}
\begin{figure}[!t]
\centering
\vspace{-0.2cm}
\includegraphics[width=\linewidth]{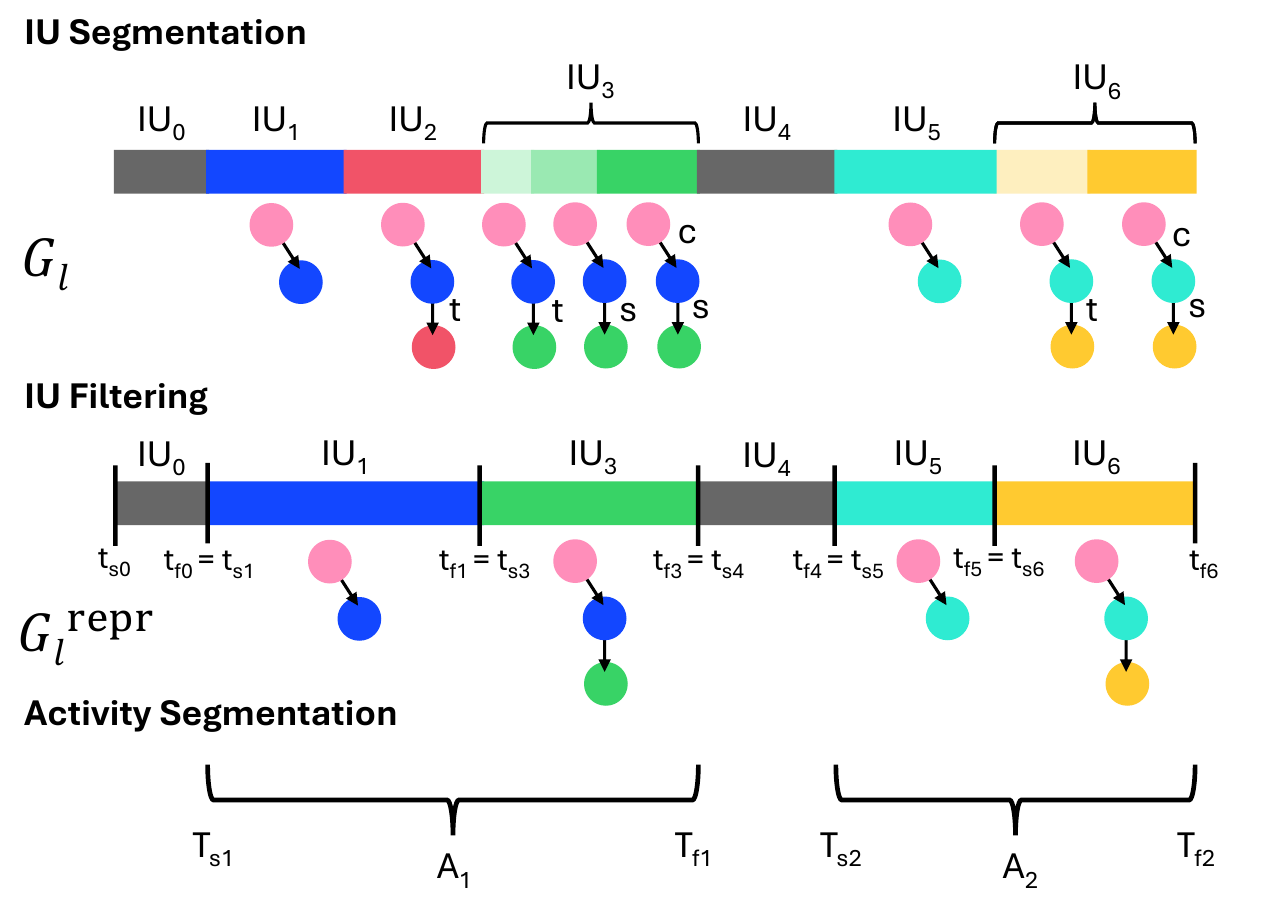}
\vspace{-0.8cm}
\caption{(top) Scene graph sequence segmentation into IUs and (mid) subsequent filtering. (bottom) Activities identification grouping IUs with same $HO$.}
\label{fig:task_segm}
\vspace{-0.5cm}
\end{figure}

The task segmentation pipeline is summarized into Algorithm \ref{algo:task_segm}.
Firstly, IU segments are created by grouping consecutive similar scene graphs, i.e., with the same topology and whose nodes have the same IDs.
The top half of Fig. \ref{fig:task_segm} illustrates the scene graph sequence segmentation into IUs. A new segment is created each time the scene graph $G[k]$ associated with frame $k$ has a new topology (transition between $\text{IU}_1$ and $\text{IU}_2$) or involves new object IDs (transition between $\text{IU}_2$ and $\text{IU}_3$) with respect to the previous frame $k-1$. However, if the interaction type changes but the graph topology and the involved objects remain unchanged, no segmentation occurs (as seen in the three green segments belonging to the same $\text{IU}_3$, where $HO$ turns to \textit{contact-only} and $OO$ evolves into \textit{significant}). Consecutive frames for which a graph was not created are grouped into a single IU (as for $\text{IU}_0$), indicating a period of no interactions.
We distinguish between two types of IUs: an IU$_{HO}$ denotes a temporal segment where only a $HO$ occurred, while an IU$_{HOO}$ is a temporal segment where both a $HO$ and an $OO$ occurred. 

\alglanguage{pseudocode}
\begin{algorithm} [b!]
\caption{Task Segmentation} \label{algo:task_segm}
\begin{algorithmic}[1]
\State $IU \gets \{\}$ \Comment{\emph{IU Segmentation}}
\State $l = 0$
\For{$k \in$ [$0$, $t_{\text{end}}$]}
    \If{$G[k]$ is NOT similar to $G[k-1]$}
        \State $l \gets l + 1$ { \small \Comment{\emph{G[k] starts a new IU}}}       
    \EndIf
    \State Add $G[k]$ to $IU[l]$
\EndFor

\For{$l \in$ [$0$, $\text{tot}_{\text{IU}}$]} \Comment{\emph{IU Filtering}}
    \If{$IU[l]$ is \textit{temporary}}
        \State Merge $IU[l]$ with the adjacent $IU_{HO}$
    \EndIf
\EndFor
\For{$l \in$ [$0$, $\text{tot}_{\text{IU}}$]} \Comment{\emph{Extract Representative Graph}}
    \State $G_{l}^\text{repr} \gets \text{Eq. } (\ref{eq:g_repr})$
\EndFor
\State $A \gets \{\}$ \Comment{\emph{Activity Segmentation}}
\State $n = 0$
\For{$l \in$ [$0$, $\text{tot}_{\text{IU}}$]}
    \If{$G^{\text{repr}}_{l}$ and $G^{\text{repr}}_{l-1}$ encode different $HO$}
        \State $n \gets n + 1$ {\small \Comment{\emph{IU[l] starts a new Activity}}}
    \EndIf   
        \State Add $IU[l]$ to $A[n]$
\EndFor
\end{algorithmic}
\end{algorithm}

Following IUs segmentation, a filtering process is executed to address temporary IUs, those composed by graphs containing a \textit{temporary} $OO$ which do not evolve into \textit{significant}. As these interactions are not strictly required for task completion, such IU$_{HOO}$s are treated as IU$_{HO}$s.
This is illustrated by $\text{IU}_2$ in Fig. \ref{fig:task_segm} (mid), which becomes a part of $\text{IU}_1$. 
As a result, the $l$-th IU is characterized by its temporal boundaries $[t_{s_l}, t_{f_l}]$ and by a representative graph $G_l^\text{repr}$ defined as:
\begin{equation}
\label{eq:g_repr}
    G_l^\text{repr} = \underset{G[k], \, \forall\, \text{frame $k$ in [$t_{s_l}$, $t_{f_l}$]}}{\arg\min} MI_{ho}[k].
\end{equation}
This means that among all the scene graphs generated during the execution of the $l$-th IU within the time interval $[t_{s_l}, t_{f_l}]$, we select the one whose $MI$ between the hand and the object (stored in the corresponding edge) is minimum. 
This way, in the case of an IU$_{HO}$, $G^\text{repr}_{HO}$ describes the instant when the manipulation begins. While for an IU$_{HOO}$, $G^\text{repr}_{HOO}$ marks when a stable relative configuration is assumed by the two objects, such as their final pose at the end of a pick-and-place task.
Moreover, for each IU$_{HOO}$, we evaluate the $MI_{ho}$ trend to determine whether the movement was intended merely to reposition $o_m$ or if it involved a more complex interaction with the background object. To achieve this, we check if $MI_{ho}$ is monotone decreasing. If it is, it means that the hand just finished the manipulation close to the background object and a new attribute called \textit{Interaction Complexity} for the $OO$ edge in $G^\text{repr}_{HOO}$ is set to false. Otherwise, the attribute is set to true, indicating a more complex interaction.
The rationale behind selecting this specific graph will become evident in Subsection \ref{subsec:plan_gen}. For now, it is important for the reader to note that $G_l^\text{repr}$ encodes the graph topology, the IDs of the nodes in the scene graphs belonging to the $l$-th IU, and the poses that the represented objects assumed in frame $k^\text{repr}$.

Finally, by grouping IUs that share the same $HO$, we can identify and discern the activities of the task, as illustrated in Fig. \ref{fig:task_segm} (bottom). In $\text{A}_1$, the hand interacts with the object depicted by the blue node, while in $\text{A}_2$, with the cyan-colored one.
The $n$-th \textit{activity} $A_n$ is defined by its temporal boundaries $[T_{s_n}, T_{f_n}]$ and by the representative graphs of the included IUs.

\vspace{-0.2cm}
\subsection{Primitives Extraction}
Once each individual activity is isolated, we can examine the sequence of IUs that define it. 
Note that we assume hand movements and manipulation of objects are the sole causes of changes in the environment and, consequently, of variations in IUs. Therefore, two adjacent IUs can be respectively considered as the preconditions and effects of specific hand behaviors. 

To automatically identify environmental changes and determine the human hand action causing them, we undertake two steps: 
\begin{inparaenum}[(a)]
    \item comparing representative graphs for the two adjacent IUs and 
    \item mapping the result into high-level primitives.
\end{inparaenum}

To compare representative scene graphs we define a difference operation between $^\text{eff}G^\text{repr}$, representing the IU describing the effects of the hand action and $^\text{prec}G^\text{repr}$ representing the IU indicating its preconditions:
\begin{equation}
    ^\text{diff}G^\text{repr} = \, 
        ^\text{eff}G^\text{repr} - \, 
        ^\text{prec}G^\text{repr}.
\end{equation}
This difference returns the subgraph describing the interaction that has started (positive difference) or the one that has expired (negative difference). 
Thus, this operation yields four fundamental results:
\begin{enumerate}[i)]
    \item  $^\text{eff}G^\text{repr}_{HO}$, denoting the start of a new $HO$; 
    \item  $- \, ^\text{prec}G^\text{repr}_{HO}$, denoting the end of a $HO$;
    \item  $^\text{eff}G^\text{repr}_{OO}$, denoting the start of a new $OO$;
    \item  $- \, ^\text{prec}G^\text{repr}_{OO}$, denoting the end of an $OO$.
\end{enumerate}
where $G^\text{repr}_{HO}$ and $G^\text{repr}_{OO}$ encode the $HO$ and the $OO$ features, respectively.
Since we assume that set of objects moving as a unity by the hand cannot detach by themselves and we are currently focusing on changes in the environment caused by a single hand, results involving variations of $OO$ can only pertain to \textit{static} $OO$ and never \textit{dynamic} $OO$.  

The subsequent step involves inferring which hand behavior is responsible for each transition and mapping it to one or a combination of high-level primitives. In this study, we considered three primitives \textendash \texttt{move}, \texttt{grasp}, and \texttt{release} \textendash as they can describe all simple actions performed by human hands \cite{guha2013minimalist}.
Given the in-hand object $o_m$ and the background object $o_j$, the fundamental $^\text{diff}G^\text{repr}$ listed earlier are mapped as follows (also clarified by pseudo-code in Algorithm \ref{algo:primitives_extr}):
\begin{enumerate}[i)]
   \item \texttt{move} through $o_m$ and \texttt{grasp} $o_m$. 
   If a new $HO$ began, it means that the hand reached such an object, grasped it, and has started to manipulate it;
   \item \texttt{release} $o_m$. If a $HO$ ended, it indicates that the object is no longer in-hand;
   \item \texttt{move} through $o_j$.
   If a new \textit{static} $OO$ began, it implies that the hand has reached the background object;
   \item continue with the next instruction, or \texttt{move} to a custom pose. If a static $OO$ ended, it suggests that the hand has moved away from the background object.
\end{enumerate}

\alglanguage{pseudocode}
\begin{algorithm} [!t]
\caption{Primitives Extraction} \label{algo:primitives_extr}
\begin{algorithmic}[1]
\State P $\gets \{\}$ {\small \Comment{\emph{list of primitives for A[n]}}}
\For{$l \in$ [$1$, $\text{tot}_{IU \in A[n]}$]}
    \State $^\text{prec}G^{\text{repr}} \gets G^{\text{repr}}_{l-1}$ {\small \Comment{\emph{$G^{\text{repr}}$ of hand action preconditions}}}
    \State $^\text{eff}G^{\text{repr}} \gets G^{\text{repr}}_{l}$ {\small \Comment{\emph{$G^{\text{repr}}$ of hand action effects}}}
    \State $^\text{diff}G^\text{repr} \gets ^\text{eff}G^{\text{repr}} - ^\text{prec}G^{\text{repr}}$   
    \If{$^\text{diff}G^\text{repr} == ^\text{eff}G^{\text{repr}}_{HO}$}
        \State Add \texttt{move} to P {\small \Comment{\emph{a new HO starts}}}
        \State Add \texttt{grasp} to P
    \ElsIf{$^\text{diff}G^\text{repr} == -^\text{prec}G^{\text{repr}}_{HO}$}
        \State Add \texttt{release} to P {\small \Comment{\emph{a HO ends}}}
    \ElsIf{$^\text{diff}G^\text{repr} == ^\text{eff}G^{\text{repr}}_{OO}$}
        \State Add \texttt{move} to P {\small \Comment{\emph{a new OO starts}}}
    \ElsIf{$^\text{diff}G^\text{repr} == -^\text{prec}G^{\text{repr}}_{OO}$}
        \State Add \texttt{move} to P {\small \Comment{\emph{a OO ends}}}
    \EndIf
\EndFor
\end{algorithmic}
\end{algorithm}

\begin{figure}[t!]
\centering
\vspace{-0.1cm}
\includegraphics[width=\linewidth]{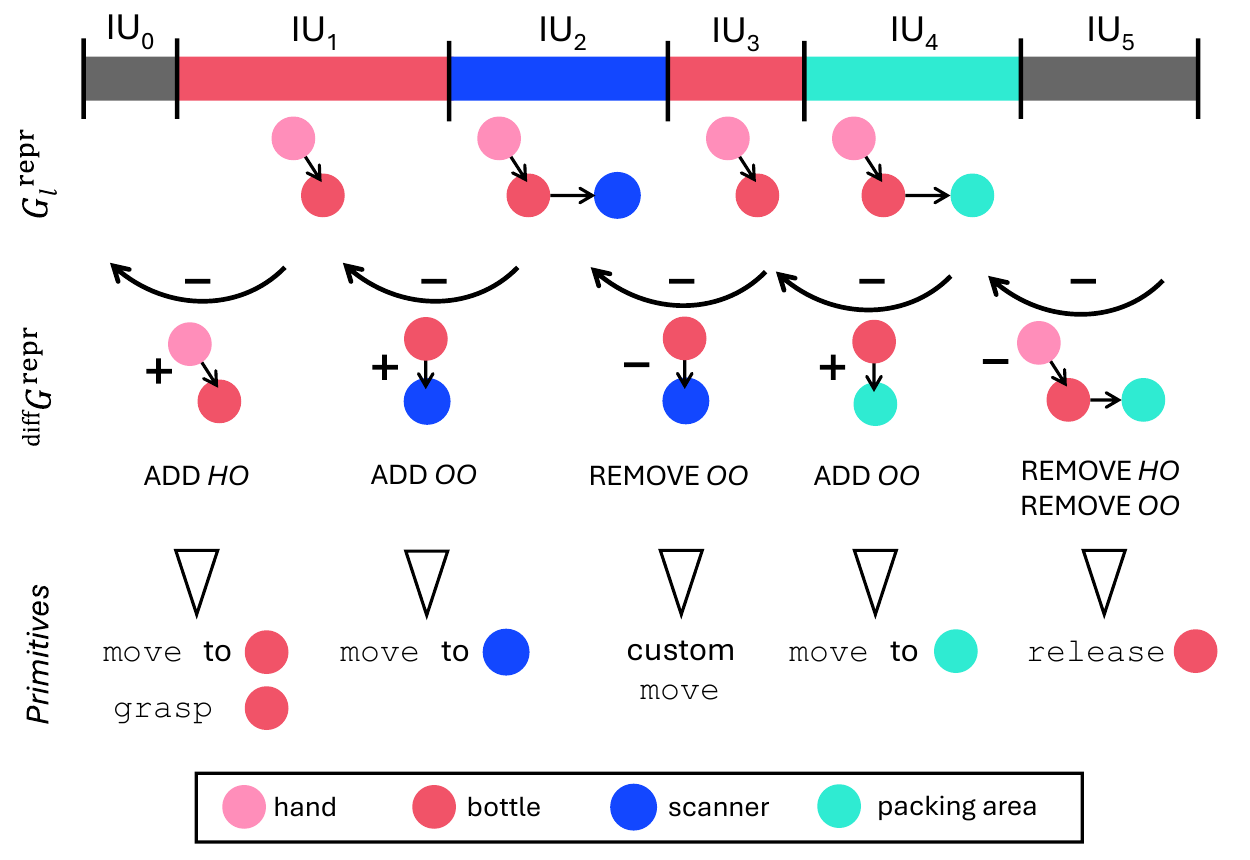}
\vspace{-0.6cm}
\caption{Extraction of motion primitives from inspection of IUs sequence within an activity.}
\label{fig:primitives}
\vspace{-0.2cm}
\end{figure}

To clarify the primitive extraction procedure, let's consider the example of a cashier working at a checkout counter. In this scenario, the cashier's task involves grabbing items from a conveyor belt, scanning them to determine the price, and placing them in the packing area. The resulting sequence of IUs and the retrieved $G^\text{repr}$ are shown in Fig. \ref{fig:primitives} (top). 
The hand starts manipulating an object, e.g. an oil bottle ($\text{IU}_1$), which is brought into proximity with the scanner but is not placed on it ($\text{IU}_2$). Then, a new interaction occurs only between the hand and the bottle ($\text{IU}_3$), persisting until the bottle is deposited in the packing area, generating a new $OO$ ($\text{IU}_4$). 
Comparing $G^\text{repr}$ in pairs, we compute the sequence of $^\text{diff}G^\text{repr}$ and determine the motion primitives for the robot replica (see Fig. \ref{fig:primitives} (mid and bottom)).
The beginning of a $HO$ with the bottle is mapped into a \texttt{move} primitive toward the bottle and \texttt{grasp} primitive of the bottle. The beginning of an $OO$ with the scanner returns again a \texttt{move} primitive to reach the scanner. Contrarily, the end of the $OO$ could not require any mapped action for the robot. 
Subsequently, when the bottle starts a \textit{static} $OO$ with the packing area, a \texttt{move} primitive toward the packing area is added. Lastly, the conclusion of the $HO$ is mapped into the \texttt{release} primitive of the manipulated bottle.

\vspace{-0.4cm}
\subsection{Plan Generation}
\label{subsec:plan_gen}

\begin{figure}
\centering
\vspace{-0.2cm}
\includegraphics[width=0.8\linewidth]{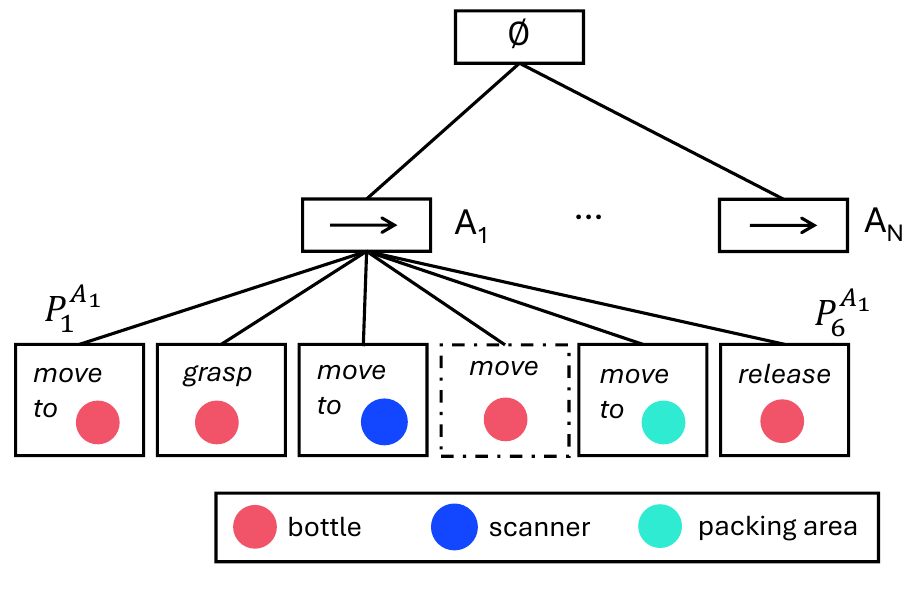}
\vspace{-0.4cm}
\caption{Automatically generated Behavior Tree where \textit{sequence} nodes represent the activities $\text{A}_1, ...\text{A}_N$ demonstrated within the task while the \textit{action} nodes correspond to the primitives $\text{P}_1, ...\text{P}_M$ within the respective activity.}
\label{fig:bt_ex}
\vspace{-0.5cm}
\end{figure}

A BT is a rooted tree structure used in robotics to model execution plans. It consists of internal nodes for control flow and leaf nodes for action execution or condition evaluation. The \textit{root} node sends a signal, called \textit{tick}, to propagate through the tree and trigger the execution of its children. Children return statuses such as SUCCESS, FAILURE, or RUNNING to indicate the outcome of their execution. 
Given their modular structure, BTs align effectively with the hierarchical task description derived from our analysis. 
Specifically, we translate each activity into a subtree where a \textit{sequence} control node serves as the root, and the primitives that constitute the activity are represented as \textit{action} execution children nodes within this subtree.
This way, we obtain a BT consisting of subtrees, each corresponding to an identified activity within the task and containing a number of \textit{action children} equal to the primitives composing that activity.
With this configuration, each \textit{sequence} node will return a SUCCESS status if and only if all the included \textit{action} nodes are executed without errors and in the correct order.
Considering again the scenario of the cashier, the resulting BT takes on the structure depicted in Fig. \ref{fig:bt_ex}. The activity of pricing the oil bottle, denoted as $\text{A}_1$, is represented by a \textit{sequence} node with six children (i.e., \textit{action} nodes $\text{P}_1^{\text{A}_1}$ to $\text{P}_6^{\text{A}_1}$), reflecting the identified primitives. Other activities demonstrated by the cashier ($\text{A}_2$ to $\text{A}_N$, e.g., pricing other items or moving the separator) are mapped into separate subtrees.

\begin{figure}
\centering
\includegraphics[width=0.7\linewidth]{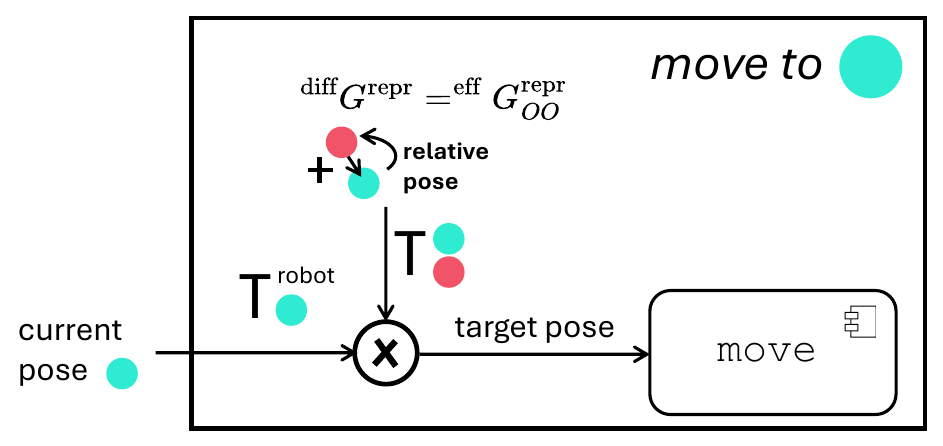}
\vspace{-0.3cm}
\caption{Action node \textit{move to packing area}.}
\vspace{-0.5cm}
\label{fig:move_node}
\end{figure}

The \texttt{grasp} and \texttt{release} primitives are included in \textit{action} nodes responsible for providing commands to the robot's gripper for closing and opening, respectively. The \texttt{move} primitive, on the other hand, when incorporated into a \textit{move to $o_\text{target}$} \textit{action} node, has a slightly more complex structure. Instructing the robot on how to reach $o_\text{target}$ requires knowledge of 
\begin{inparaenum}[(i)]
    \item the relative poses assumed by $h$ and $o_m$ or by $o_m$ and $o_j$ during the demonstration and
    \item the current pose of the objects within the robot's workspace at the replication time.
\end{inparaenum} 
The first information is extracted from $^\text{diff}G^\text{repr}$ and embedded within the \textit{action} node in the shape of a transformation matrix between the two elements ($T^{o_m}_{h}$ or $T^{o_j}_{o_m}$, respectively representing the pose of the reference frame attached to the hand $h$ w.r.t. to the one attached to the manipulated object $o_m$, and that of $o_m$ w.r.t. the one attached to the background object $o_j$). 
The second information is instead measured and received as input by the BT.
For example, let's consider the \textit{move to packing area} action node from the cashier example, shown in Fig. \ref{fig:move_node}. It embeds $T^\text{packing area}_\text{bottle}$, indicating the pose of the bottle with respect to the packing area when such a pose became stable, i.e., just before the bottle was released. 
This node receives as input the current pose of the packing area relative to the robot, which is translated into the transformation matrix $T^{\text{robot}}_{\text{packing area}}$. The end effector's target pose is obtained by multiplying the two matrices.  
This is the pose that the end effector should reach in order for the manipulated bottle to assume the observed pose with respect to the packing area. Subsequently, the \texttt{move} primitive plans and executes a trajectory to reach this desired end effector pose.
Note that the planned trajectory may vary based on the object's characteristics, which are known and can be incorporated into the graph structure. Thus, the trajectory could include grasping, transporting and placing the object for pick-and-place tasks, or in the case of a non-prehensile object, it might be shifted to the desired destination.
If $^\text{diff}G^\text{repr}$ reports the \textit{Interaction Complexity} between $o_m$ and $o_j$ set to true, the algorithm maps $o_m$'s motion into two action nodes: the first is a \textit{move to} node, which handles reaching $o_j$, while the second one is responsible for executing the specific motion pattern required for $o_m$, learned apart. In the cashier example, the bottle is not simply located on the scanner, so the interaction movement should be learned. In Fig. \ref{fig:bt_ex}, the \textit{move bottle} action node, outlined with a dashed border, is designed to possibly incorporate the learned scanning motion.

With this design, we can automatically generate the robot's execution plan from a human demonstration.
Moreover, by providing the current poses of the involved objects, the robot replica can generalize to various scenarios beyond the demonstrated one while maintaining the same BT structure.

\vspace{-0.3cm}
\section{Experiments}
\vspace{-0.1cm}
Our method evaluation comprised two experiments\footnote{The multimedia attachment related to the conducted experiments can be found in the additional materials provided with the article submission and also online at \href{https://www.youtube.com/watch?v=NgCo9np0zAI}{https://www.youtube.com/watch?v=NgCo9np0zAI}.}. The experimental campaign aimed first to assess the functionality of the video processing and task representation, leading to the automatic generation of a robot execution plan. Subsequently, we evaluated the generalization of the generated plan to various environmental conditions. 
Before proposing the multi-subject experiment, we conducted a pilot study to demonstrate the validity of the trends of entropy-based metrics in segmenting and describing manipulation tasks and compare the outcomes with a velocity-based segmentation pipeline. 
We asked a participant to manipulate a pan by moving it towards themselves, rotating it to stir the contents, and then returning it to its original position.
\vspace{-0.8cm}
\subsection{Task Representation Assessment (Exp 1)}

\subsubsection{Experimental Setup}

To evaluate our pipeline, we conducted a multi-subject experiment involving $10$ participants, comprising $5$ males and $5$ females with an average age of $28.4 \pm 2.4$ years. Among them, $8$ were right-handed and $2$ were left-handed\footnote{The entire experimental process was conducted at the Human Robot Interfaces and Interaction (HRII) Lab, Istituto Italiano di Tecnologia (IIT), in compliance with the principles of the Declaration of Helsinki. The protocol received approval from the ethics committee of Azienda Sanitaria Locale (ASL) Genovese N.3 under Protocol IIT\_HRII\_ERGOLEAN 156/2020.}. 
The experimental setup involved an RGB camera (Intel RealSense D435i) positioned in a top-down (bird's eye) view, with the image plane aligned parallel to the working plane as shown in Fig. \ref{fig:setup}(a).
To enable robust detection of the $6$D pose of objects and hands we used a marker-based detection system. ArUco markers were attached to the back of the hand and strategically positioned on the objects, preserving natural movements during manipulation\footnote{A file containing objects' specifics along with corresponding ArUco positions is included in the open-source documentation at \href{https://doi.org/10.5281/zenodo.13846970}{HANDSOME}.}. It is worth noting that marker-less objects or hand detectors are continuously improving \cite{shan2020understanding, wu2024general}, presenting opportunities for future integration into our framework.
Given that we analyzed tasks which primarily involved larger movements performed on the plane, calculations regarding spatial relationships and entropy exchange were conducted considering $x$ and $y$ components of positional signals. However, the poses saved as attributes for each graph node include both position and orientation in $3$D space. 
Furthermore, we set the image acquisition frequency at $30$ Hz and selected a window size $w=1.3$ s covering $40$ samples to calculate the average distance and entropy measures. This means that at each instant $t$, we evaluated the positions of hands and objects approximately half a second before and after $t$. This window size provided a suitable estimate for the time it takes for a human to alter the relationships between objects \cite{ziaeetabar2018recognition}. Although setting $w$ is a crucial step, opting for a value between $1$ and $2$ seconds is appropriate for capturing interaction occurrences and obtaining a high-level representation of the task.
Entropy measures were calculated using position data quantized with $q=1$ cm. We set $\epsilon_{MI}=0.05$, the minimum distance between hand and object to $d^{th}_{ho} = 0.15$ m, while the minimum distance between two objects to $d^{th}_{oo} = 0.2$ m, reflecting object identification based on marker position. Distance thresholds among scene elements were defined with reference to \cite{diehl2021automated}.  
Finally, to assess how a signal changed over time, we examined the values taken in the most recent $20$ ($= w/2$) measurements. The signal's derivative was considered negative if the majority of differences between consecutive values of the signal exhibited a decreasing trend.
We utilized an Alienware laptop equipped with an Intel Core i7 processor, an NVIDIA RTX 2080 GPU, and 32 GB of RAM.
The architecture was implemented in Python and ran on Ubuntu 20.04. The functions used to compute entropy and MI are part of \href{https://dit.readthedocs.io/en/latest/measures/shannon.html}{\texttt{dit}} Python package specifically designed for discrete information theory \cite{dit}.

\subsubsection{Experimental Protocol}
Participants were instructed to perform manual tasks, with half of them operating within a kitchen context and the remaining in a workshop context.
Each subject completed $5$ different tasks, repeated $5$ times, involving object manipulation, resulting in a total of $250$ executions recorded\footnote{The entire documentation about HANDSOME dataset \cite{merlo2024hand} is available as open-source and can be accessed at \href{https://doi.org/10.5281/zenodo.13846970}{\textbf{HANDSOME}}.}. 
The tasks included: 
\begin{inparaenum} [(i)]
    \item simple manipulation to analyze the interaction between only the hand and manipulated objects;
    \item manipulation of assembled objects, 
    simple pick-and-place actions, and 
    more complex activities to examine various types of interactions between objects.
\end{inparaenum}
Table \ref{tab:tasks} reports the tasks performed in the kitchen context and their equivalent in the workshop context. 

\subsubsection{Method Validation}
First, a video demonstration of Task $1$ was exploited to analyze the trend of $MI_{ho}$ between the hand and an object combined with their average relative distance $\overline{d}_{ho}$ over time to capture features involved in the generation of a $HO$ and its evolution into \textit{contact-only} $HO$. Similarly, we examined the behavior of $MI_{oo}$ and average distance $\overline{d}_{oo}$ of two objects involved in a \textit{dynamic} $OO$, from a demonstration of Task $2$. The detection of \textit{static} $OO$ was tested in Task $3$, Task $4$, and Task $5$, considering how the trend of $H(\overline{d}_{oo})$ determined the discernment between \textit{temporary} and \textit{significant} $OO$. However, we present results related to one performance of Task $4$ and Task $5$, which exhibit more articulated activities. 
Subsequently, we evaluated the accuracy of the segmentation process across all executions of Tasks $3$, Task $4$, and Task $5$ by comparing each segmentation output with a manually generated reference (ground truth) and considering the BT plan derived from it.
Given the influence of the scene encoding quality on segmentation and the fact that the correctness and feasibility of the generated robot plan are directly tied to the segmentation results, the segmentation assessment effectively reflects the success rate of the entire pipeline. 

\begin{table}[!t]
\caption{Manual Activities in our Dataset}
\label{tab:tasks}
\centering
\begin{tabular}{|c||c|c|}
\hline
 & \textbf{Kitchen} & \textbf{Workshop}\\
\hline
\textbf{Task 1} & move a cup & move a profile\\
\hline
\textbf{Task 2} & move a tray with two cups & move an assembly\\
\hline
\textbf{Task 3} & toast & mark a profile\\
\hline
\textbf{Task 4} & put cups on a tray & weigh and box a profile\\
\hline
\textbf{Task 5} & clean a cooker & polish a stone\\
\hline
\end{tabular}
\vspace{-0.4cm}
\end{table}

\begin{figure}
\centering
\includegraphics[width=\linewidth]{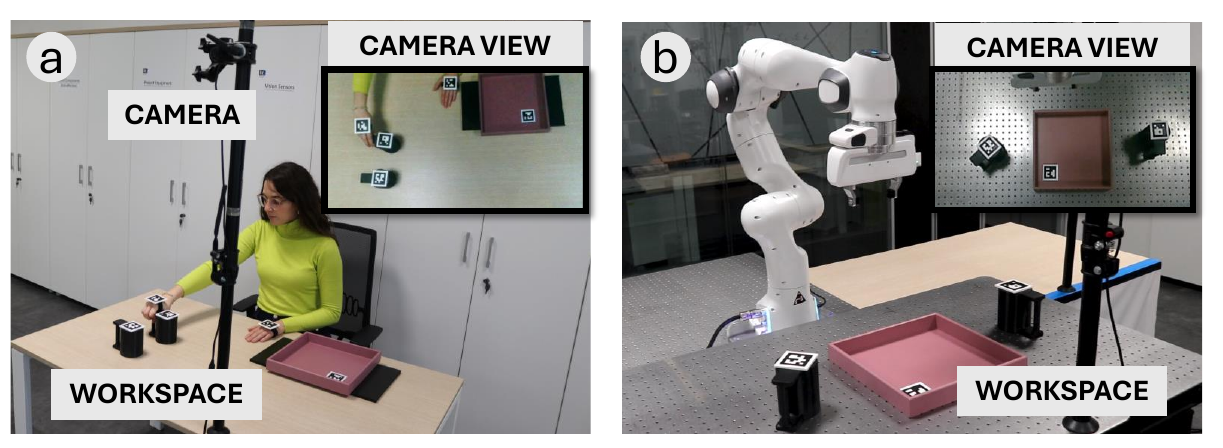}
\vspace{-0.5cm}
\caption{(a) Experimental setup and camera view during human demonstrations. (b) Robot workspace and camera view at the time of the replica.}
\label{fig:setup}
\vspace{-0.4cm}
\end{figure}

\vspace{-0.3cm}
\subsection{Plan Generalization Assessment (Exp 2)}

\subsubsection{Experimental Setup}
In the second experiment, the setup involved a Franka Emika Panda robot controlled at $1$ kHz and equipped with the Franka Hand, an electrical two-finger parallel gripper, as shown in Fig. \ref{fig:setup}(b). To determine the initial pose of objects at the beginning of the execution, the robot used data streamed by the RGB camera (Intel RealSense D435i) mounted again in a top-down view.

\subsubsection{Experimental Protocol}
We assessed the robot's ability to generalize its performance and successfully complete the required task, even under altered environmental conditions. To achieve this, we used two videos, one for each context (kitchen and workshop), of Task $4$ which is the only task composed of two activities. 
In this experiment, we assumed that only the robot could manipulate objects within the scene. As a result, the first activity in the BT was constructed using the initial objects' poses captured by the camera, while for modeling the second activity, the objects' poses were updated based on the movements adopted by the robot to complete the first one.
Since the considered tasks were designed to avoid hands rotations that could obscure ArUco markers and hindering their detection, rotations around vertical axis were only treated in the robot replica.
Note that although we predefined the grasping points on cups and profiles, exploiting the information in the scene graphs, their definition could potentially be automated.

\subsubsection{Method Validation}
First, we evaluated the case in which the objects involved in the task assumed different initial positions compared to those observed in the human demonstration. Since all information encoded within the scene graph structure pertains to the relative poses between objects, we expected the robot to place the objects in a manner that maintained these relative relations as demonstrated by the human, regardless of the initial conditions. To demonstrate this, we focused on the task of placing two cups on a tray, varying the initial pose of the cups and tray three times. 
Moreover, we verified the successful execution of the task even when objects not engaged in any significant interaction were removed from the background. For this purpose, we utilized the demonstration of weighing a profile and placing it in the box, removing one of the profiles already present in the box.

\vspace{-0.3cm}
\section{Results}

\subsection{Task Representation Assessment (Exp 1)}

\subsubsection{Comparison between Entropy and Velocity Metrics for Interaction Segmentation}
\begin{figure}
    \centering
    \includegraphics[width=1\linewidth]{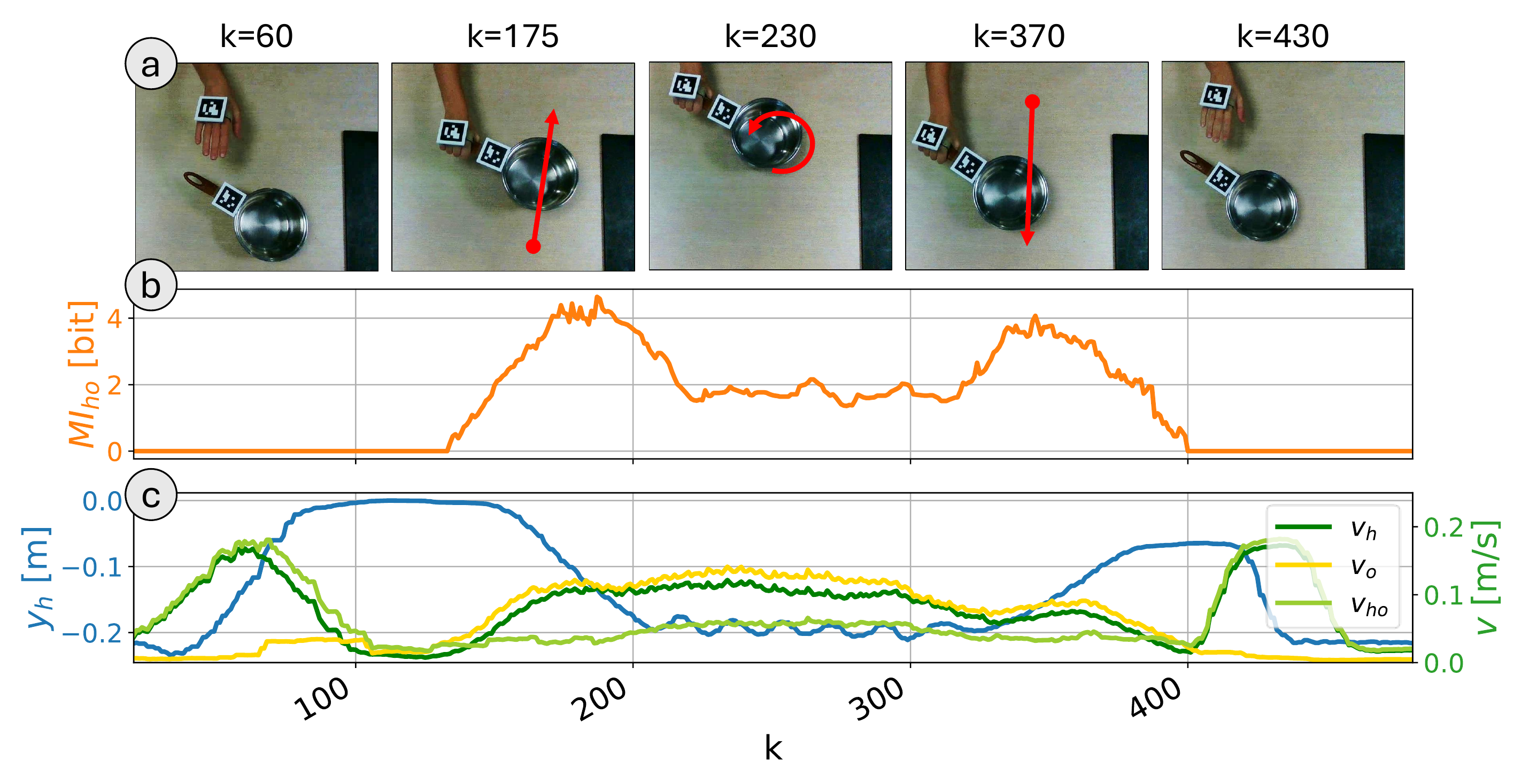}
    \vspace{-1cm}
    \caption{(a) Keyframes of the pilot experiment.
(b) Trends of hand-pan mutual information (orange line) over time. 
(c) Trends of the hand position component $y$ (blue line), of the hand speed (dark green line), of the pan speed (yellow line), and of the relative hand-pan speed (light green line).}
\vspace{-0.5cm}
    \label{fig:vel-MI}
\end{figure}

Fig. \ref{fig:vel-MI}(a) shows keyframes during pan manipulation with red arrows indicating the movement directions. The $MI_{ho}$ signal in Fig. \ref{fig:vel-MI}(b) highlights the entire manipulation period with values greater than zero, indicating that the hand and pan are moving together as a unit. In contrast, using velocity-based methods to identify this interaction would require first checking if the relative velocity $v_{ho}$ (light green line in Fig. \ref{fig:vel-MI}(c)) is near zero and then confirming that both elements are actually in motion (in dark green the hand speed $v_h$, while in yellow the pan speed $v_o$), necessitating the setting of one or more thresholds that depend on the perception system's quality. 
Moreover, we observe how $MI_{ho}$ can indicate unexpected movements during manipulation, providing insights into joint motion evolution. In Fig. \ref{fig:vel-MI}(b), peaks at $k=175$ and $k=370$ correspond to straight-line movements with significant positional changes, reflected in the blue line representing the $y$ component of hand position. During the stirring phase (around $k=230$), $MI_{ho}$ stays relatively low and constant, indicating repetitive and predictable movements. In contrast, $v_h$ remains nearly constant, making it difficult to distinguish between patterns without separately analyzing each velocity component, complicating the computation.
Additionally, we investigated the effect of adding synthetic Gaussian noise $n \sim \mathcal{N}(\mu, \sigma^2)$ to the positional data, with $\mu = 0$ and $\sigma = 0.03$ m and $0.05$ m. The velocity signals became corrupted, while $MI_{ho}$ remained more robust, still allowing for manipulation phase segmentation.

\subsubsection{Scene Representation}
\begin{figure}
\centering
\includegraphics[width=\linewidth]{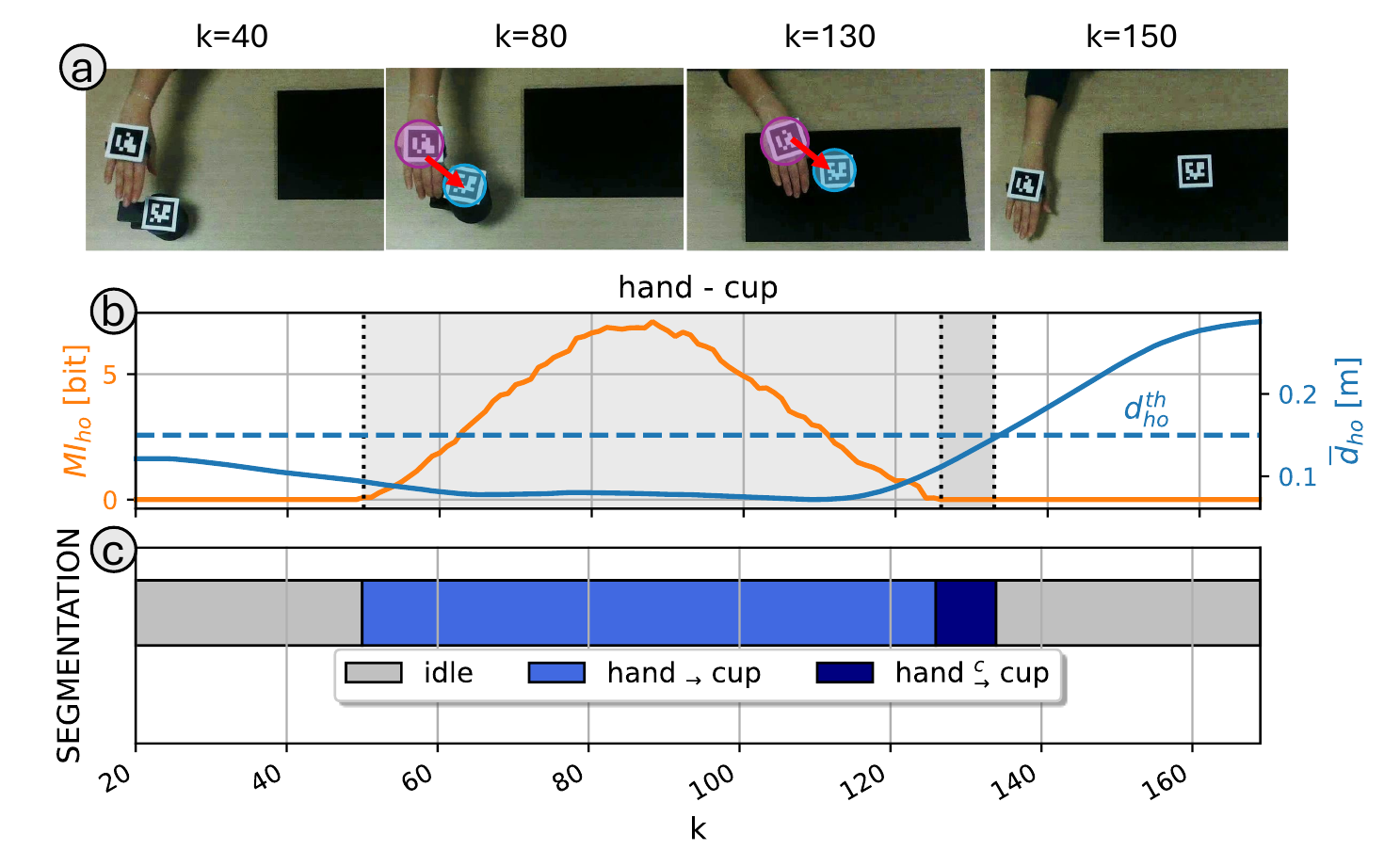}
\vspace{-0.9cm}
\caption{(a) Keyframes of a demonstration of Task $1$ in kitchen context executed by subject $3$. Scene graphs generated during manipulation and release phases include hand and cup nodes connected by edges indicating \textit{manipulation} and \textit{contact-only} $HO$, respectively. 
(b) Trends of average relative distance between the hand and the cup
(blue line) and computed mutual information 
(orange line) with a sliding time window. 
Light gray area indicates manipulation phase with stable grasp. Darker gray area shows \textit{contact-only} $HO$ during cup release. 
(c) Segmentation result, indicating time intervals that share equal graph topology, node IDs, and interaction type.
}
\label{fig:ho_result}
\vspace{-0.4cm}
\end{figure}

Considering $HO$ detection, Fig. \ref{fig:ho_result} presents the results of video processing for a participant demonstrating Task $1$, with only one object in the environment. 
During the hand-approaching phase, $\overline{d}_{ho}$ (blue curve) decreases below $d_{ho}^\text{th}$ and remains approximately constant throughout the manipulation, indicating a firm grasp of the object.
In the manipulation phase (highlighted with a light gray shaded area), $MI_{ho}$ (orange curve) surpasses the threshold $\epsilon_{MI}$ and exhibits a bell-shaped profile due to the window shift over time along the hand and object positional data. In the other phases (approaching, grasping, and distancing), $MI_{ho}$ is instead below the threshold $\epsilon_{MI}$.
Note that proximity between the hand and the cup is a necessary but not a sufficient condition to detect an interaction. Indeed, no scene graph was generated for the initial frames, as can be seen in frame $k=40$ of Fig. \ref{fig:ho_result} (a).
During the manipulation phase, the scene graphs assumed the topology presented in frame $k=80$, indicating an $HO$ between the hand and the cup.
This interaction did not cease when $MI_{ho}$ approached zero. Instead, as long as the average relative distance in the sliding time window remained below the minimum threshold, a \textit{contact-only} $HO$ persisted (highlighted by the darker gray shaded area). The graph topology, shown for frame $k=130$, remained the same while the interaction type changed from \textit{manipulation} to \textit{contact-only} $HO$. Fig. \ref{fig:ho_result} (c) illustrates the segmentation result, indicating with color-coded blocks the graphs that share the same topology, node identities, and interaction types. 
Note that shades of the same color indicate different interaction types, e.g., segments representing the \textit{manipulation} (blue) and the \textit{contact-only} (dark blue) $HO$. Subsequently, as the hand moved away, no graph was generated until the end of the video.

\begin{figure}[t!]
\centering
\includegraphics[width=\linewidth]{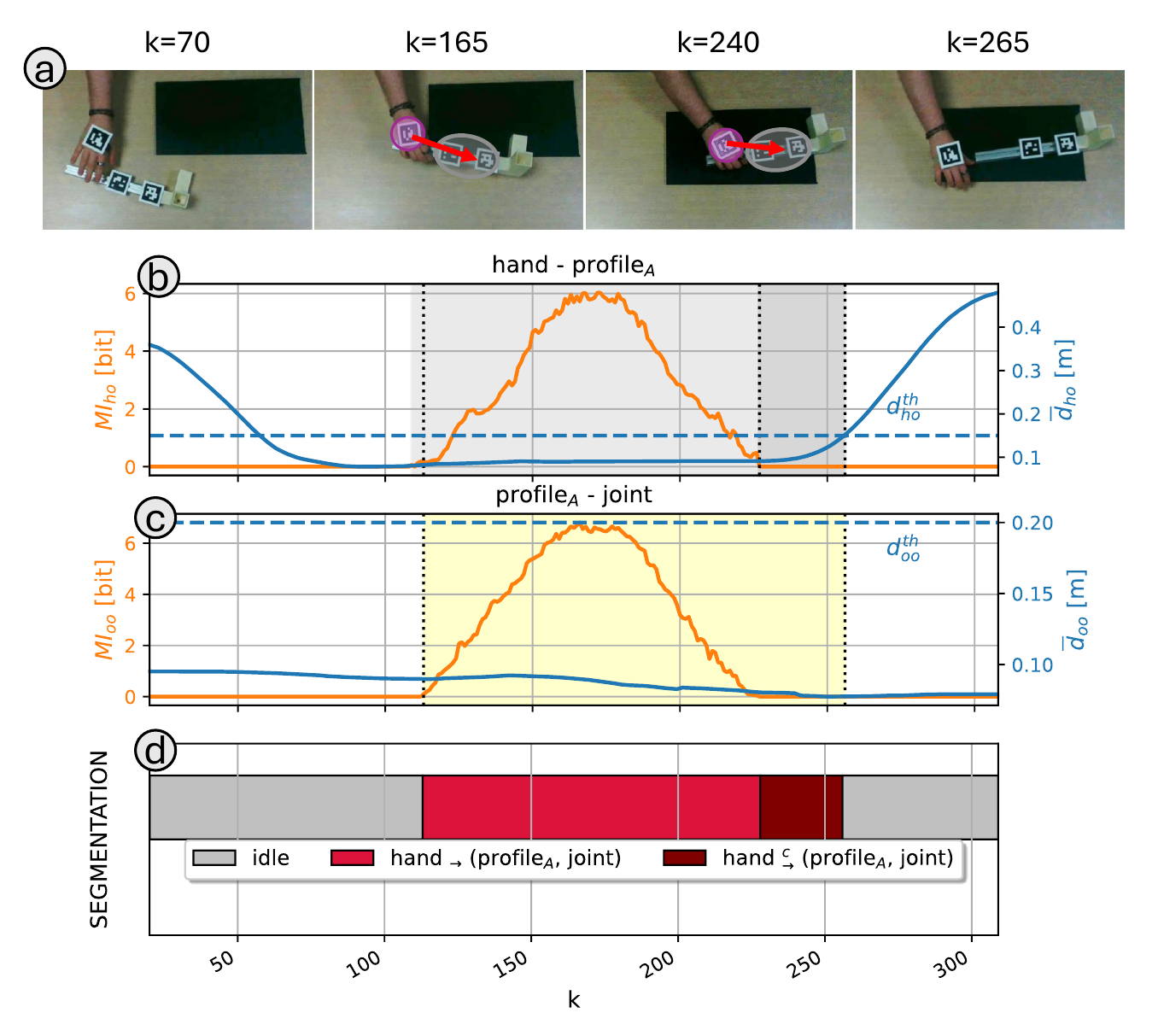}
\vspace{-0.9cm}
\caption{(a) Keyframes and generated scene graphs during a demonstration of Task $2$ in workshop context executed by subject $6$. A moving unity composed of profile$_A$ and joint nodes was detected. 
(b) Trends of average relative distance and of MI between hand and profile.
(c) Trends of average relative distance and of MI between profile and joint within a sliding time window.
A yellow shaded area highlights the period during which a \textit{dynamic} $OO$ was detected. (d) Segmentation result.}
\label{fig:dOO_result}
\vspace{-0.6cm}
\end{figure}

The manipulation of an assembly is instead reported in Fig. \ref{fig:dOO_result}. In Task $2$, the participant manipulated an aluminum profile (profile$_A$) fitted into an angle joint. The algorithm identified a moving unity $u_m$, as observed from scene graph topology at $k=165$ and $k=240$, since the average relative distance profile$_A$-joint $\overline{d}_{oo}$ (blue curve) remains stable below $d_{oo}^\text{th}$ (dashed blue curve), and their $MI_{oo}$ (orange curve) is non-zero during the manipulation phase (see Fig. \ref{fig:dOO_result} (c)). The generated scene graphs present a \textit{dynamic} $OO$ between the two objects in the assembly for the period highlighted by the yellow shaded area, which lasts for the phases characterized by \textit{manipulation} and \textit{contact-only} $HO$ (i.e., light and dark gray areas in Fig. \ref{fig:dOO_result} (b)). This suggests that the interaction between the hand and the assembly persisted after manipulation but changed its type, as shown by the segmentation results in Fig. \ref{fig:dOO_result} (d).

\begin{figure*}[!t]
\centering
\includegraphics[width=\linewidth]{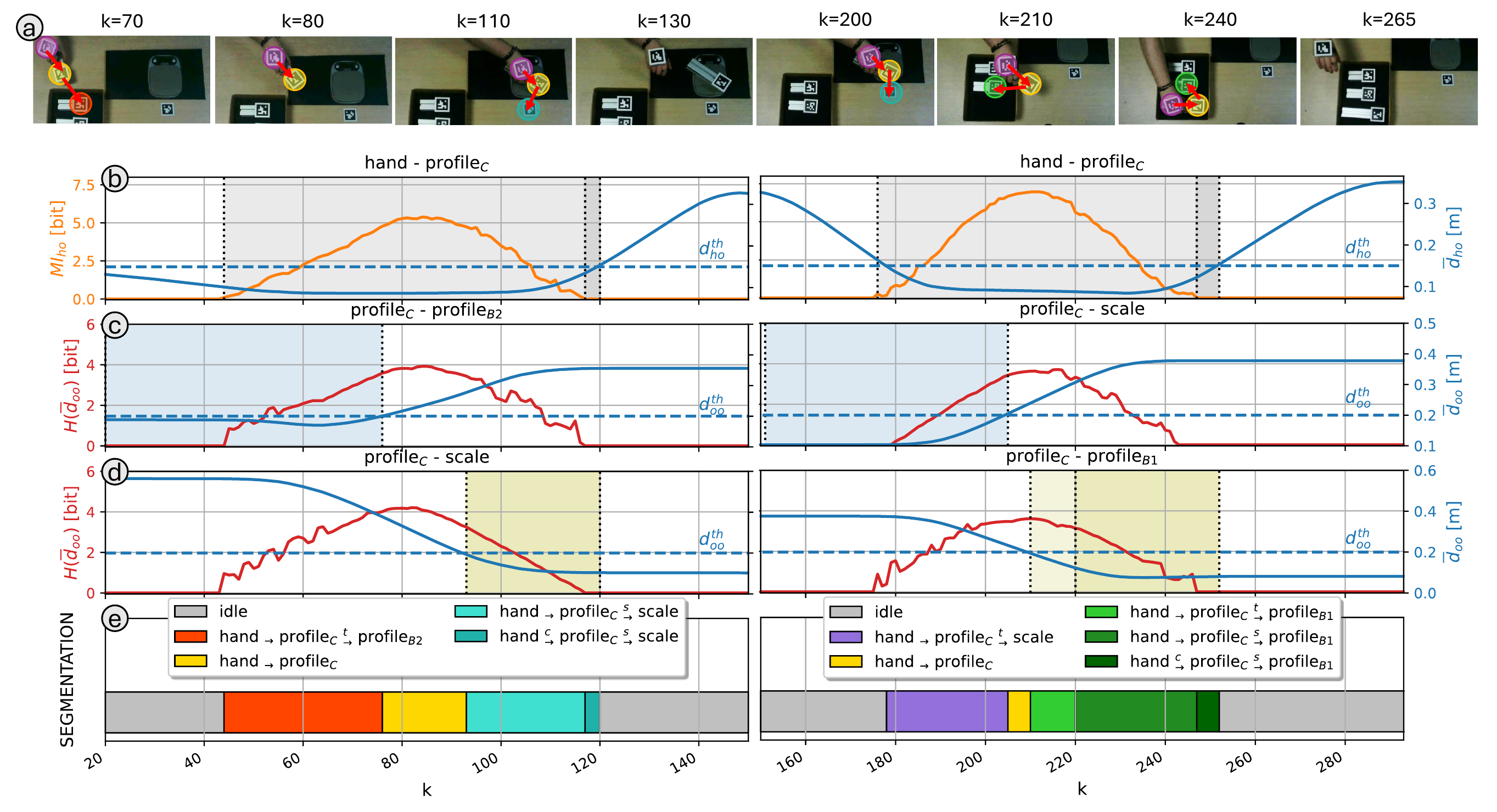} 
\vspace{-0.8cm}
\caption{(a) Keyframes of a demonstration of Task $4$ executed by subject $6$ in workshop context. Generated scene graphs encode interactions between the hand and profile$_C$, as well as potential interactions with other objects in the background. 
(b) Trends of average relative distance between hand and manipulated object and of their MI are depicted, along with 
(c) trends of average relative distance between manipulated and background objects and of the entropy of such a distance, indicating \textit{temporary} $OO$ and 
(d) \textit{significant} $OO$. 
(e) Segmentation result.
}
\label{fig:sOO_result}
\vspace{-0.3cm}
\end{figure*}

\begin{figure}[b!]
\centering
\vspace{-0.7cm}
\includegraphics[width=\linewidth]{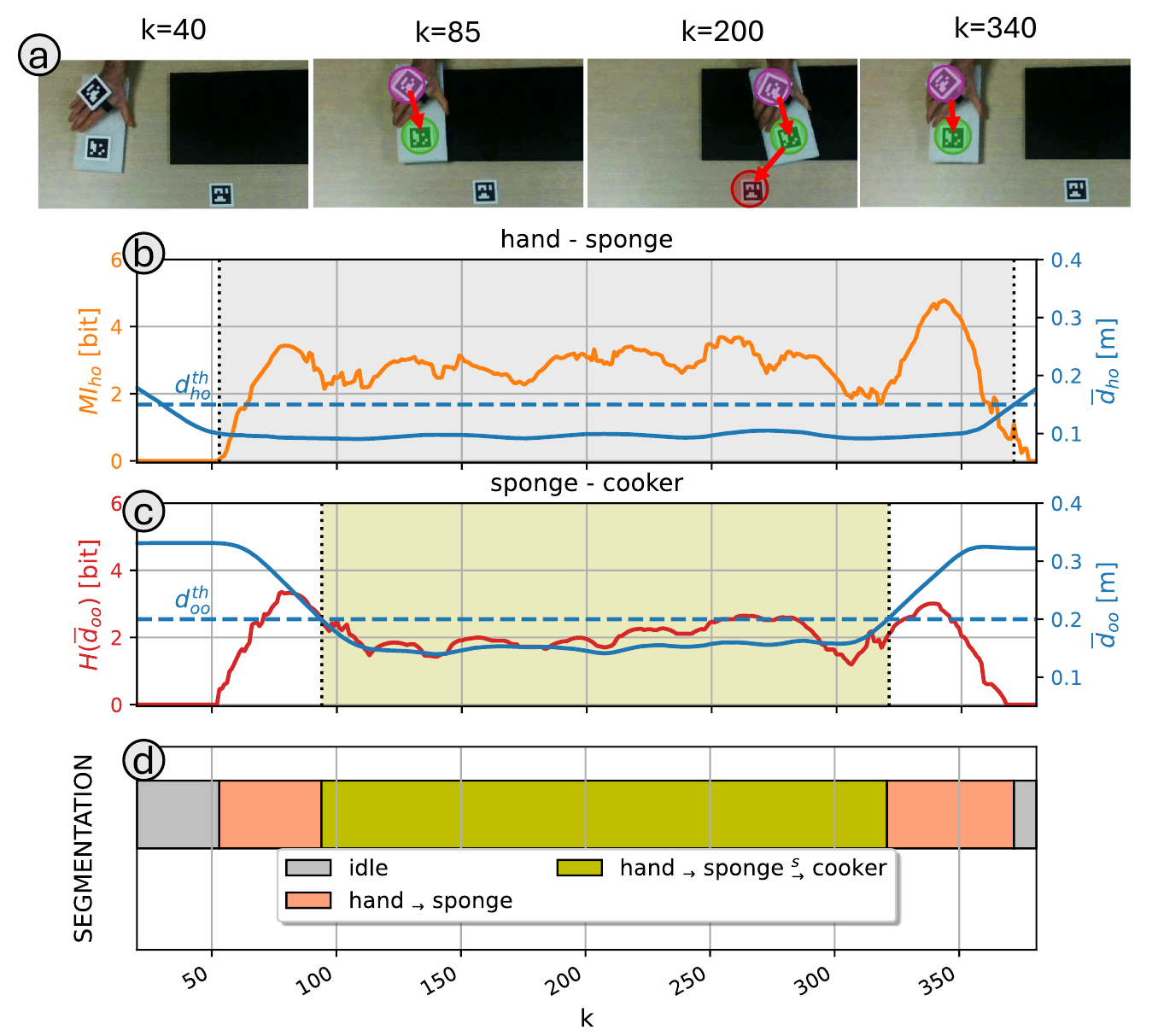}
\vspace{-0.8cm}
\caption{(a) Keyframes of a demonstration of Task $5$ in kitchen context executed by subject $1$. Scene graphs generated encoding either only $HO$ between hand and sponge or also $OO$ between sponge and cooker. (b) Trends of average relative distance between hand and sponge and of their MI.
(c) Trends of average relative distance between sponge and cooker and of the entropy of such a distance.
(d) Segmentation result.
}
\label{fig:sOO_clean_result}
\end{figure}

\begin{figure}[!b]
\centering
\vspace{-0.4cm}
\includegraphics[width=\linewidth]{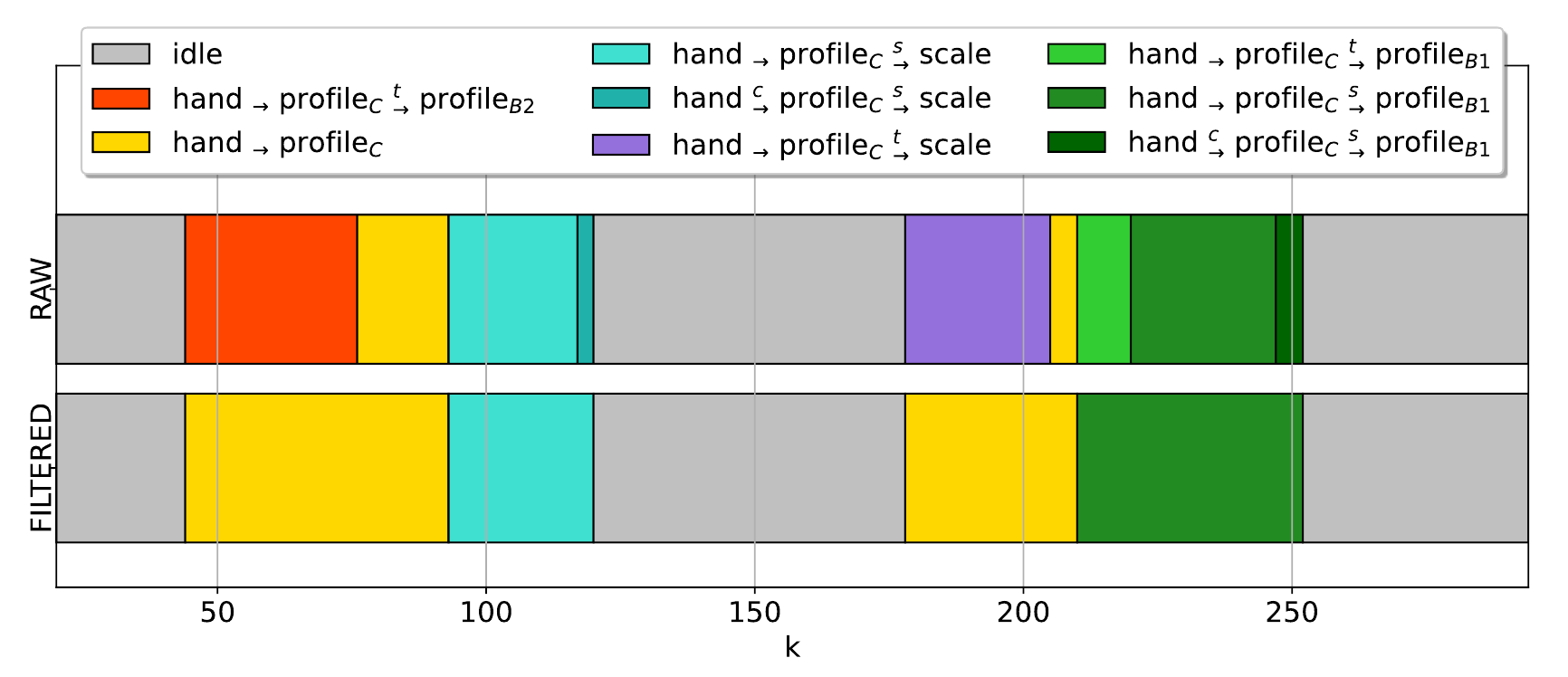}
\vspace{-0.8cm}
\caption{Segmentation process for a demonstration of Task $4$ in workshop context executed by subject $6$. The first \textit{temporary} $OO$ is included in the adjacent $HO$, as well as the \textit{temporary} $OO$ between profile$_C$ and scale in the second activity. The two cyan blocks are unified indicating the same $OO$ as well as the three green blocks.} 
\label{fig:seg_soo}
\end{figure}
\begin{figure*}[!b]
\centering
\includegraphics[width=\linewidth]{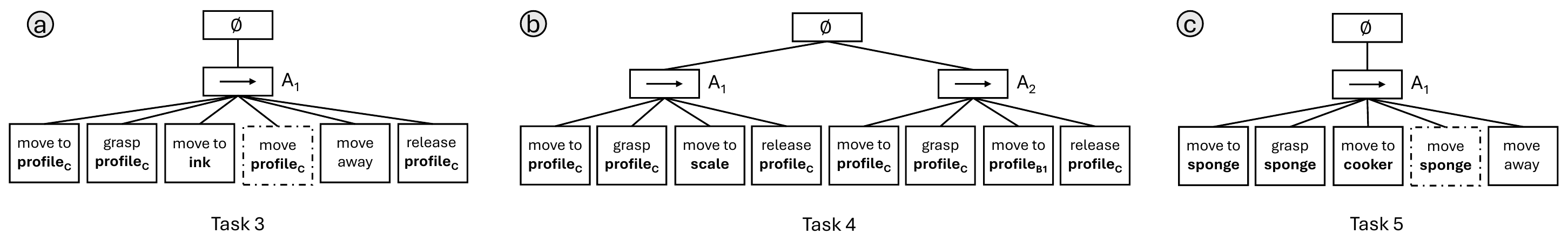}
\vspace{-0.8cm}
\caption{Generated Behavior Trees for Task 3 (a), Task 4 (b), and Task 5 (c).}
\label{fig:3bt}
\end{figure*}
\begin{figure*}[!b]
\centering
\includegraphics[width=\linewidth]{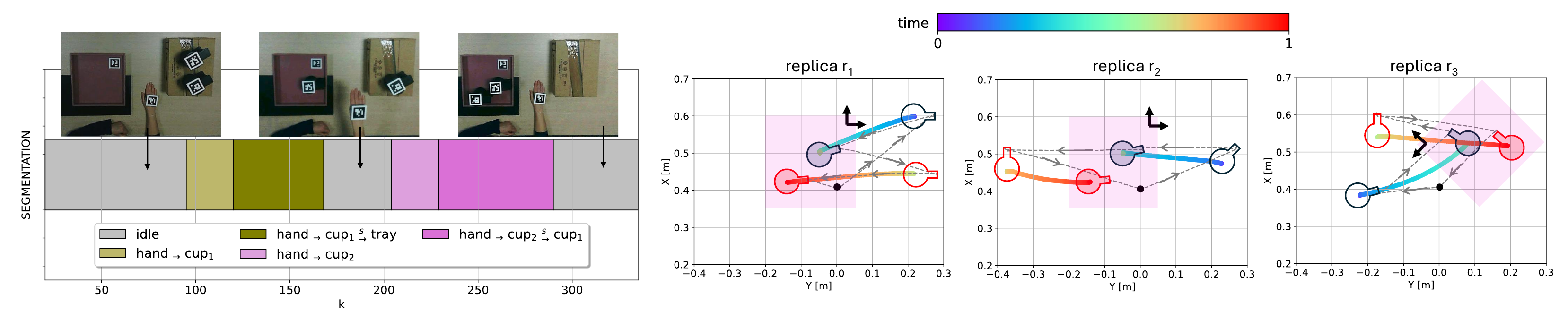}
\vspace{-0.6cm}
\caption{(left) Segmentation of a Task $4$ demonstration in a kitchen context and (right) robot replication results in three different configurations. The pink square indicates the tray and cup$_1$ and cup$_2$ are colored in blue and red respectively. The black dot indicates the end effector's homing position, dashed gray lines depict its trajectory, and colored trails show cups' trajectories over time.}
\label{fig:expe2cups}
\vspace{-0.3cm}
\end{figure*}

In the demonstration of Task $4$ reported in Fig. \ref{fig:sOO_result}, the participant performed the weighing and boxing of profile$_C$ in an environment with other objects visible in the background. 
During the manipulation of profile$_C$, the latter remained involved in a static $OO$ with profile$_{B2}$ inside the box, due to their $\overline{d}_{oo}$ being under threshold (as shown for $k=70$). However, this $OO$ was labeled as \textit{temporary}, since the entropy trend $H(\overline{d}_{oo})$ was increasing. 
This indicates that, in this particular demonstration, encoding the interaction with profile$_{B2}$ may not be necessary for the task's correct execution. It is worth noting that in other demonstrations where the initial pose of profile$_C$ differed slightly, the hand tended to follow the edge of the box, treating it as an obstacle to be avoided in order to reach the scale. Consequently, the $OO$ was considered \textit{significant} in those cases.
This $OO$ was maintained until the two profiles were sufficiently distant, as highlighted by the blue shaded area on the left of Fig. \ref{fig:sOO_result} (c), at which point a change in graph topology was registered ($k=80$).
As the scale was approached, a static \textit{significant} $OO$ was established between the profile and the scale ($k=110$). This raised from the concurrent decreasing trend of $H(\overline{d}_{oo})$, depicted by the red curve in Fig. \ref{fig:sOO_result} (left, d). 
During the weighing, no graph was generated since the hand moved away from profile$_C$.
Upon re-grasping it, a \textit{temporary} $OO$ was observed with the scale, resulting from their initially close proximity, which quickly increased as indicated by the trend in $H(\overline{d}_{oo})$. The graph topology shown for $k=200$ persisted until profile$_C$ moved sufficiently away from the scale.
When profile$_{B1}$, already inside the box, was approached, an $OO$ was detected between profile$_C$ and profile$_{B1}$. Initially classified as \textit{temporary}, this interaction became \textit{significant} when $H(\overline{d}_{oo})$ between the two profiles started to decrease. This transition was highlighted by the beginning of the darker yellow area in Fig. \ref{fig:sOO_result} (right, d). However, the graph topology remained the same as shown in $k=210$ and $k=240$.
Colored blocks at the bottom of Fig. \ref{fig:sOO_result} (both left and right) illustrate the evolution of scene graphs over time.

The detection of \textit{static} $OO$ generated in case of a more complex hand motion is illustrated in Fig. \ref{fig:sOO_clean_result}.
The subject used a sponge to clean an induction cooker, moving the sponge back-and-forth three times over the surface. As a result, the pattern of $MI_{ho}$ (orange curve) exhibited a sequence of peaks and valleys (see Fig. \ref{fig:sOO_clean_result} (b)). Each valley corresponds to a pronounced change in motion direction; the deceleration that occurs during the direction change causes the time window $w$ to include more similar positional values, leading entropy to decrease.
In Fig. \ref{fig:sOO_clean_result} (c), $H(\overline{d}_{oo})$ (red curve) presents two peaks at the start and end of the sponge manipulation, indicating when the sponge approached and retreated from the cooker. During cleaning, an $OO$ is observed between the sponge and the cooker, as indicated by the graph topology at time $k=200$. 
The oscillating trend of $MI_{ho}$ indicates that this $OO$ involves unpredictable motion resulting in the \textit{Interaction Complexity} attribute being set to true.
This $OO$ is labeled as \textit{significant} due to the decrease in $H(\overline{d}_{oo})$ at the onset of the interaction, and it persisted until the sponge moved away from the cooker. During the wiping phase, $H(\overline{d}_{oo})$ reflects the fluctuations of the distance between the two objects.

\subsubsection{Mapping to Robot Instructions}
The assessment of segmentation performance across the $150$ executions of Task $3$, $4$, and $5$ revealed a success rate of $92$\%. 
However, errors were primarily attributed to incorrect scene graph generation, due to perception issues (constituting $4$\% of cases). Noisy marker position signals resulted in increased entropy, causing incorrect detection of $HO$ and dynamic $OO$. The remaining $4$\% of errors was related to how the subjects performed their motions, especially in Task $4$ within the workshop context. 
Fig. \ref{fig:seg_soo} illustrates the segmentation pipeline conducted on the scene graphs generated for the task demonstration showcased in Fig. \ref{fig:sOO_result}.
Two activities, each consisting of two IUs, have been identified, separated by an idle period. 
However, in a few executions, a single activity was obtained, since after placing the profile on the scale, the hand did not move away far enough, resulting in the hand never releasing it.
In the first activity, the $OO$ with profile$_{B2}$ resulted \textit{temporary} and therefore discarded, while in a minority of executions, such $OO$ turned \textit{significant}. 
The BT shown in Fig. \ref{fig:3bt} (b) originates from Task $4$ segmentation presented in Fig. \ref{fig:seg_soo}. The two activities are mapped to two subtrees.
Action nodes are similar for the two activities: \textit{move to} profile$_C$ and \textit{grasp} it, \textit{move it to} the background object, and \textit{release} it.
Segmentation errors influence BT generation and, thus, the robot's replica. For instance, if \textit{temporary} $OO$ with profile$_{B2}$ is not discarded, an additional primitive is generated, instructing the robot to move towards profile$_{B2}$ before weighing profile$_C$, even if it is not needed. Moreover, this would prevent the weighing if profile$_{B2}$ was not in the environment.
On the other hand, if the two activities are not separated, the action node for releasing profile$_C$ on the scale is not generated.
Fig. \ref{fig:3bt} (a) shows the BT generated to instruct the robot about how to mark profile$_C$ with ink. Due to the identification of a unique activity, the BT presents just one sequence node: among its children we found the \textit{move profile$_C$} node which is configured to execute the profile marking operation.
Then, the custom target pose for the \textit{move away} node can be selected to bring profile$_C$ close to the table surface since it is followed by the \textit{release} node.
In Fig. \ref{fig:3bt} (c), we reported the BT derived from segmenting the cooker cleaning task. The robot is instructed to grasp the sponge, reach the cooker, and to reproduce the cleaning movement if it is known.

\begin{figure} [t!]
\vspace{-0.4cm}
\centering
\includegraphics[scale=0.4]{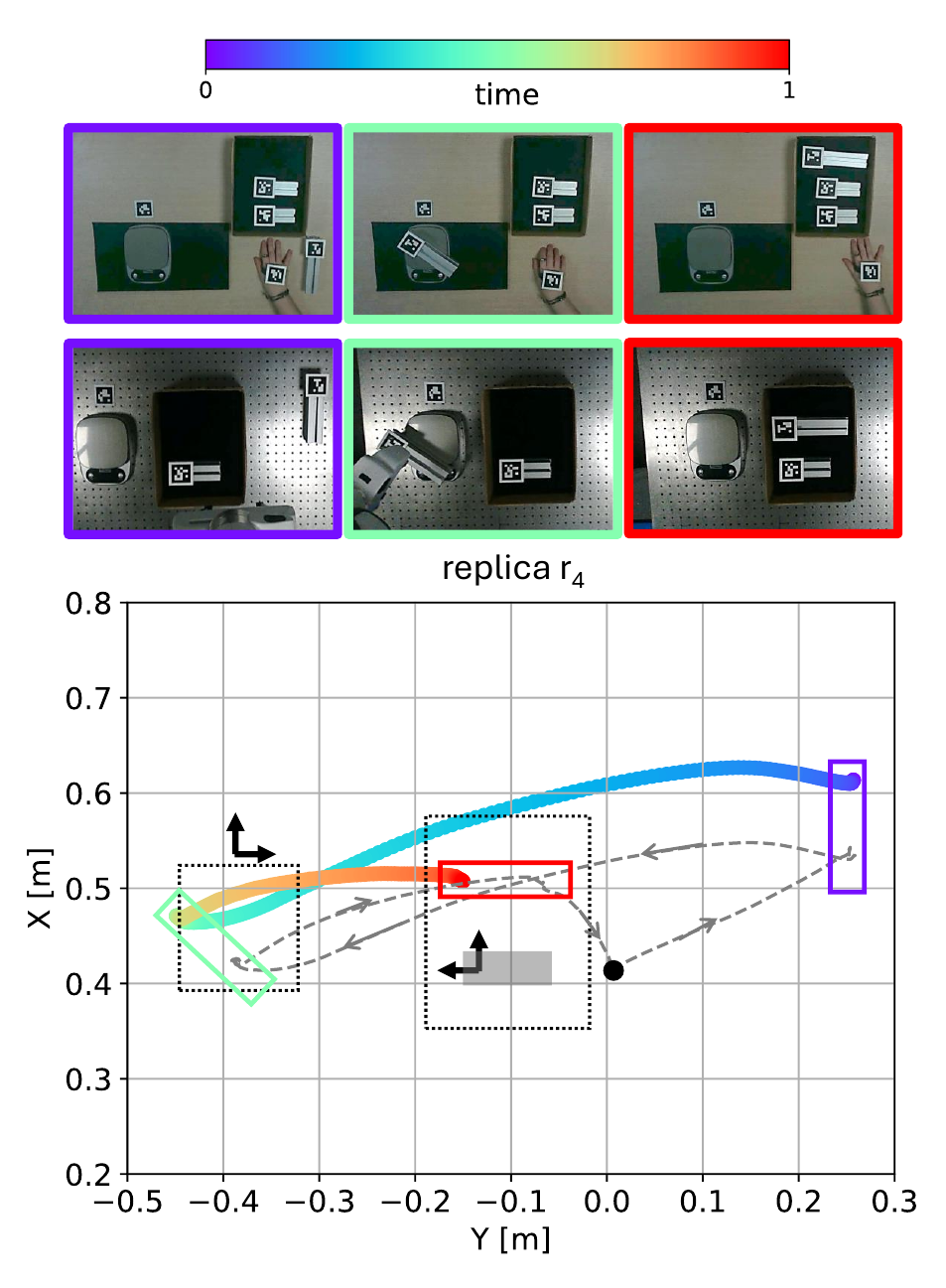}
\vspace{-0.2cm}
\caption{Robot replication of Task $4$ within a workshop context. (top) Frame contour colors denoting objects' configurations at the beginning (purple), end of the first activity (aquamarine), and end of the task (red), for both human demonstration and robot replication. (bottom) Plot representing end effector (gray dashed line) and profile$_C$ (colored trails) trajectories. Contour color of profile$_C$ indicates its position, which changes over time.}
\label{fig:expe2weigh}
\vspace{-0.3cm}
\end{figure}

\vspace{-0.2cm}
\subsection{Plan Generalization Assessment (Exp 2)}
Fig. \ref{fig:expe2cups} (left) shows the segmentation of a demonstration of Task $4$ in the kitchen context: the subject grasped cup$_1$ and placed it on the tray, repeating the operation for cup$_2$.
Fig. \ref{fig:expe2cups} (right) shows the results when the robot replicated the task in the three different arranged configurations. 
In replica $r_1$, objects assume poses similar to those observed: the tray was in the center of the workspace, and the two cups aligned in a column on the right. 
In $r_2$, we maintained as fixed the position of the tray, with cup$_1$ on the right and cup$_2$ on the left.
In $r_3$, we changed the pose of the tray as well, and both cups were positioned on the left.
For all the replications, the robot first grasped cup$_1$ (drawn in blue) and then cup$_2$ (drawn in red), as in the human performance.
In the figure, the black dot represents the end effector's homing position, and the dashed gray line depicts its trajectory executed in the direction indicated by the arrows. The colored trails show the trajectory of the two cups over time. 
It is important to note that even though the absolute configurations of the elements in the scene may change, the final relative $3$D position and the $z$ orientation between the markers on the cups and the marker on the tray remain consistent. We observed slight variations in the cup configurations compared to the reference, particularly in the orientation of the cup handles relative to the upper edge of the tray. However, since these variations are consistent across the replicas, we attribute them to limitations in the accuracy of the perception system.

Fig. \ref{fig:expe2weigh} illustrates the robot's replication of Task $4$ within a workshop context, following the BT structure depicted in Fig. \ref{fig:3bt} (b). Despite profile$_{B2}$ being absent from the scene, the robot successfully completed the weighing and boxing of profile$_C$. At the end of each activity, the relative pose between profile$_C$ and the two background objects remained consistent with the human reference.
In Fig. \ref{fig:expe2weigh} (top) the contour colors of frames indicate the configurations of the objects involved at the beginning (purple), at the end of the first activity (aquamarine), and at the end of the task (red), in both the human demonstration and the robot replica. Same colors are used to depict the profile configuration over time.

\vspace{-0.3cm}
\section{Discussion}
Results of the \textit{Task Representation Assessment} demonstrated the ability of the proposed framework to generate a compact yet comprehensive description of manual activities. Using information theory, relevant interactions between scene elements were effectively captured by a single video demonstration, and only the active part of the scene was encoded, leaving the background out. 
In particular, the use of mutual information proved beneficial in accurately detecting instances of object manipulation and the coordinated movement of two objects, e.g., in an assembly. This approach offered a more robust alternative to relying solely on contact estimation or distance measurements. Analyzing the exchange of information between scene elements helped prevent the misidentification of interactions merely based on the proximity between two elements. Our strategy allows also to capture motion dependencies directly and more effectively than analyzing velocity trends, even with less reliable perception data.
Furthermore, by examining the trend of entropy-based measures, we could identify events in hand motion patterns, helping also to distinguishing intentional interactions between objects from accidental ones, even with just a single human demonstration. 
This enables an effective representation of the executed task even when task-irrelevant objects are present. Moreover, it ensures representation consistency when the same activity is demonstrated regardless of variations in the background configuration.

The proposed segmentation pipeline adeptly identified and isolated individual activities within the task execution by exploiting scene graph properties. It should be noted that this activity separation strategy offers the potential to represent articulated tasks featuring a greater number of activities, each comprised of an arbitrary number of interaction units. 
Subsequently, from the internal arrangement of each activity, we can easily derive the preconditions and effects of human actions and exploit them to establish the mapping to robot primitives. 
Overall, the segmentation of $150$ multi-subject video demonstrations of Tasks $3$, $4$, and $5$, compared to ground truth annotations, achieved a high accuracy, reflecting the quality of the scene encoding. At the same time, accurate segmentation ensured the generation of a correct execution plan, a process that relies on the temporal sequence of the retrieved task segments. Deviations in segmentation indeed led to incorrect or unfeasible plans. Therefore, the obtained $92$\% success rate not only reflects segmentation quality but is also an indicator of a valuable pipeline’s overall performance.
The $4$\% of failures due to perception issues suggests our algorithm's reliance on quite accurate position data. However, in our current approach, which consists of a one-shot human-to-robot transfer using a single video, we demonstrated that entropy-based measures can mitigate errors in interaction detection that arise from analyzing velocity profiles, even adding synthetic Gaussian noise to the retrieved positional signals. To further enhance the pipeline's real-world applicability, applying filtering techniques and learning algorithms that use data from multiple repetitions of the same task could be beneficial.

The structure of the Behavior Trees properly mapped the task organization uncovered by our representation, using control and action nodes to define the necessary steps for the execution.
Moreover, thanks to the effective task representation, the robot was able to successfully replicate the human-demonstrated task with new environmental conditions. Indeed, the encoding of the desired relative poses between the involved objects enabled the robot to reproduce this configuration regardless of their initial poses. 
In addition, by recognizing interactions that are insignificant for task execution and excluding them from the execution plan, the robot was able to perform successfully even when background objects varied. Therefore, the robot's performance is not affected if additional objects are present in the scene or if objects that are irrelevant to the task execution are absent at the time of the robot replica. 
The method proposed in this paper proves efficient in generating machine instructions for human demonstrations of pick-and-place activities. Conversely, at the current stage, it is not handling the transfer of complex movements like those required for a cleaning or polishing task; the robot is able to reach the surface to clean or polish, but it cannot move the end effector as the human hand during demonstration. 
However, this limit could be addressed by exploiting algorithms for trajectory learning, which could be applied to hand motion data regarding the execution of the desired movement that our segmentation process can already isolate.
The learned motion can then be used to feed the \textit{move} action node we have prepared for handling complex interaction movements.

\vspace{-0.2cm}
\section{Conclusion}

This paper established the groundwork for an intuitive and cost-effective robot programming technique. Specifically, robots were enabled to learn to perform manual tasks by observing single human hand demonstrations from RGB videos, exploiting the simplicity of recording using standard RGB cameras and the abundance of web resources. Using information theory and scene graph properties, our framework can recognize and analyze the various interactions in manual tasks, extract their structure, and convert it into a robot execution plan that can generalize to varying environmental conditions. Experimental evaluation was conducted by collecting a dataset of multi-subject demonstrations of diverse manual activities involving real-world objects in realistic scenarios, achieving an overall pipeline success rate of 92\%. 
By providing the gathered dataset as open-source, we aim to offer a benchmark for similar approaches in the field.
\\
To enhance our application's performance and real-world applicability, future work could focus on advancing the perception module with vision foundation models for object segmentation and 3D reconstruction. That would enable the demonstrator to move freely and allow us to test our algorithm in handling perception data from marker-less systems, which tend to be more error-prone. Additionally, integrating learning from demonstration algorithms could also help the robot master specific movements for complex tasks like marking or cleaning. Developing methods to adapt plans based on online environmental conditions and user's preferences represents another promising avenue for improvement.
Finally, we would like to investigate the information exchange during bi-manual activities, encoding the coordination modality within our representation and retrieving a suitable mapping to either robot-robot or human-robot collaborative plans.

\vspace{-0.3cm}
\bibliographystyle{IEEEtran}
\bibliography{biblio}

\newpage

\end{document}